\documentclass[10pt,twocolumn,letterpaper]{article}

\usepackage{wacv}
\usepackage{wacv}
\usepackage{times}
\usepackage{epsfig}
\usepackage{graphicx}
\usepackage{amsmath}
\usepackage{amssymb}
\usepackage{textcomp}
\usepackage{booktabs,siunitx}
\usepackage{subfig}
\usepackage{graphicx}
\usepackage{algorithm}
\usepackage{algpseudocode}
\usepackage[toc,page]{appendix}
\usepackage{bm}
\usepackage[accsupp]{axessibility}

\def\wacvPaperID{953} % Enter the WACV Paper ID here

\wacvfinalcopy % *** Uncomment this line for the final submission

\ifwacvfinal
\fi

\ifwacvfinal
\usepackage[breaklinks=true,bookmarks=false]{hyperref}
\else
\usepackage[pagebackref=true,breaklinks=true,colorlinks,bookmarks=false]{hyperref}
\fi

\begin{document}

%%%%%%%%% TITLE
\title{Enhancing Few-Shot Image Classification with Unlabelled Examples}

\author{Peyman Bateni$^{1,6}$\thanks{Authors contributed equally}\hspace{0.06in}\thanks{Work performed while at Inverted AI} \hspace{0.01in}, Jarred Barber$^{2\hspace{0.01in}*}$\thanks{Work performed while at Charles River Analytics} \hspace{0.01in}, Jan-Willem van de Meent$^3$, Frank Wood$^{1,4,5}$\\
University of British Columbia$^1$, Amazon$^2$, Northeastern University$^3$, Inverted AI$^4$, MILA$^5$, Beam AI$^6$\\
{\tt\small pbateni@cs.ubc.ca, jarred.barber@gmail.com, j.vandemeent@northeastern.edu, fwood@cs.ubc.ca}}

\maketitle

\setlength{\belowdisplayskip}{0.2pt} \setlength{\belowdisplayshortskip}{0.2pt}
\setlength{\abovedisplayskip}{0.2pt} \setlength{\abovedisplayshortskip}{0.2pt}

\ifwacvfinal
\thispagestyle{empty}
\fi

%%%%%%%%% ABSTRACT
\begin{abstract}
   We develop a transductive meta-learning method that uses unlabelled instances to improve few-shot image classification performance. Our approach combines a regularized Mahalanobis-distance-based soft k-means clustering procedure with a modified state of the art neural adaptive feature extractor to achieve improved test-time classification accuracy using unlabelled data. We evaluate our method on transductive few-shot learning tasks, in which the goal is to jointly predict labels for query (test) examples given a set of support (training) examples. We achieve state of the art performance on the Meta-Dataset, mini-ImageNet and tiered-ImageNet benchmarks. All trained models and code have been made publicly available\footnote{Code available at \href{github.com/plai-group/simple-cnaps}{github.com/plai-group/simple-cnaps}}.
\end{abstract}
\vspace{-0.2in}

\section{Introduction}

Deep neural networks have transformed machine learning and computer vision  \cite{DBLP:journals/corr/abs-1907-09408-object-detection-survey,8441512-image-classification-survey, Hossain:2019:CSD:3303862.3295748-image-captioning-survey, guz-etal-2020-neural, Krizhevsky12_AlexNet, He15_ResNet, Redmon16_YOLO, Ren15_FasterRCNN, Goodfellow2014_GANS, Scibior2021_ITRA}, enabled in part by the development of large and diverse sets of curated training data \cite{DBLP:journals/corr/SzegedyLJSRAEVR14-inception, DBLP:journals/corr/HeZRS15-resnet, Krizhevsky:2017:ICD:3098997.3065386alexnet, vgg-paper-2014, 8441512-image-classification-survey}. However, in many image classification tasks, millions of labelled examples are not available; therefore, techniques that can achieve sufficient classification performance with few labels are required. This has motivated research on few-shot learning \cite{feyjie2020semisupervised-medical, DBLP:journals/corr/abs-1904-05046-survey-on-few-shot-learning, Wang:2019:SZL:3306498.3293318-survey-of-zero-shot-learning, bellet2013survey}, which seeks to develop methods for developing classifiers with much smaller datasets. Given a few labelled ``support'' images per class, a few-shot image classifier is expected to produce labels for a given set of unlabelled ``query'' images. Typical approaches to few-shot learning adapt a base classifier network to a new support set through various means, such as learning new class embeddings \cite{Snell17_Proto, vinyals2016matching, sung2018learning}, amortized \cite{requeima2019fast, NIPS2018_7352-tadam} or iterative \cite{DBLP:journals/corr/YosinskiCBL14-finetune} partial adaptation of the feature extractor, and complete fine-tuning of the entire network end-to-end \cite{DBLP:conf/iclr/RaviL17-meta-lstm, finn2017model}. %Most recently, \cite{bateni2019improved} achieved state of the art accuracy using Simple CNAPS, a conditional neural-adaptive feature extractor with a regularized Mahalanobis-distance-based classifier that uses high-dimensional mean and covariance estimates to produce class-clusters in the feature space.

\begin{figure}
    \centering
    \includegraphics[width=2.79in]{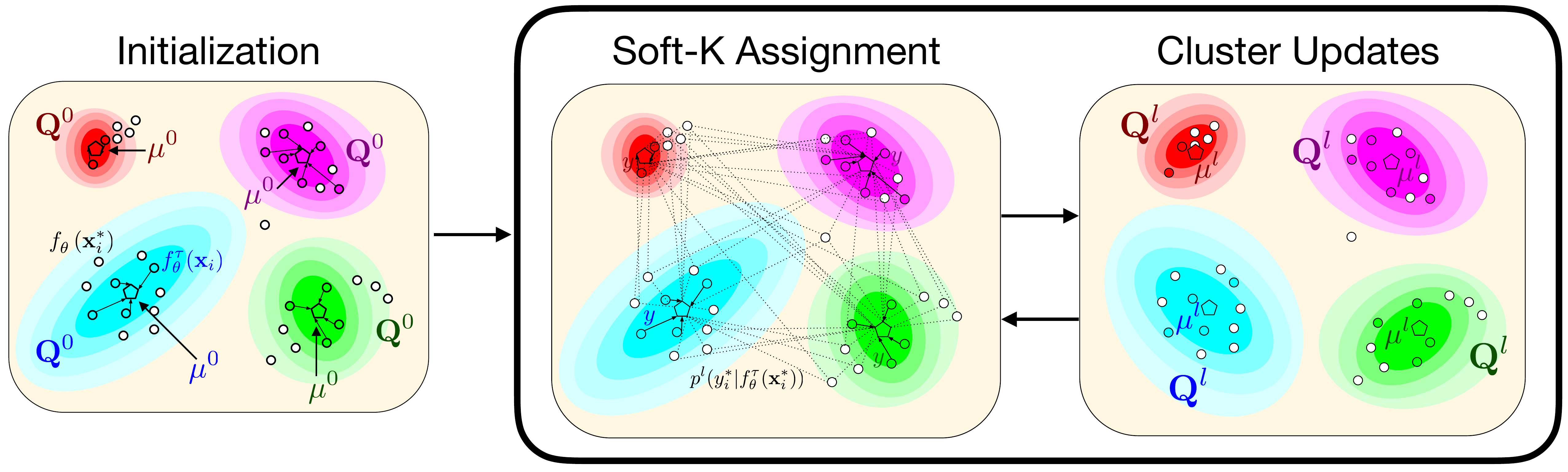}
    \vspace{-0.1in}
    \caption{Transductive CNAPS' soft k-means Mahalanobis-distance based clustering procedure. First, cluster parameters are initialized using the support examples. Then, during cluster update iterations, query examples are assigned class probabilities as soft labels and subsequently, both soft-labelled query examples and labelled support examples are used to estimate new cluster parameters.}
    \vspace{-0.2in}
    \label{fig:clustering-method}
\end{figure}

In addition to the standard fully supervised setting, techniques have been developed to exploit additional unlabeled support data (semi-supervision) \cite{DBLP:journals/corr/abs-1803-00676-tieredimagenet} as well as information present in the query set (transduction) \cite{DBLP:journals/corr/abs-1805-10002-tpn, DBLP:journals/corr/abs-1905-01436-edge-labelling-gnn}.
%While these methods only use labelled support examples for few-shot learning of visual tasks, 
%\cite{DBLP:journals/corr/abs-1803-00676-tieredimagenet}, \cite{ DBLP:journals/corr/abs-1905-01436-edge-labelling-gnn}, and \cite{ DBLP:journals/corr/abs-1805-10002-tpn} make additional use of unlabelled data, whether through a secondary support set of images without labels \cite{DBLP:journals/corr/abs-1803-00676-tieredimagenet} or by directly taking advantage of the query images, provided all at once, as unlabelled examples \cite{DBLP:journals/corr/abs-1805-10002-tpn, DBLP:journals/corr/abs-1905-01436-edge-labelling-gnn}. 
In our work, we focus on the transductive paradigm, where the entire query set is labeled at the same time. This allows us to exploit the additional unlabeled data, with the hopes of improving classification performance.  Existing transductive few-shot methods reason about unlabelled examples by performing k-means clustering with Euclidean distance \cite{DBLP:journals/corr/abs-1803-00676-tieredimagenet} or message passing in graph convolutional networks \cite{DBLP:journals/corr/abs-1805-10002-tpn, DBLP:journals/corr/abs-1905-01436-edge-labelling-gnn}. 

Since few-shot classification requires handling a varying number of classes, an important architectural choice is the final feature to class mapping. Previous methods have used the Euclidean distance \cite{DBLP:journals/corr/abs-1803-00676-tieredimagenet}, the absolute difference \cite{koch2015siamese}, cosine similarity \cite{vinyals2016matching}, linear classification \cite{finn2017model, requeima2019fast} or additional neural network layers \cite{DBLP:journals/corr/abs-1905-01436-edge-labelling-gnn, sung2018learning}. 
Bateni et al.~\cite{bateni2019improved} introduced a class-adaptive Mahalanobis distance. Their method, Simple CNAPS, uses a conditional neural-adaptive feature extractor, along with a regularized Mahalanobis-distance-based classifier. This modification to CNAPS \cite{requeima2019fast} achieved improved performance on the Meta-Dataset benchmark \cite{triantafillou2019meta}, only recently surpassed by SUR \cite{dvornik2020selecting-sur} and URT \cite{liu2020universal-urt}. 
However, its performance suffers when there are five or fewer support examples available per class.

\begin{figure*}
    \centering
    \includegraphics[width=4.85in]{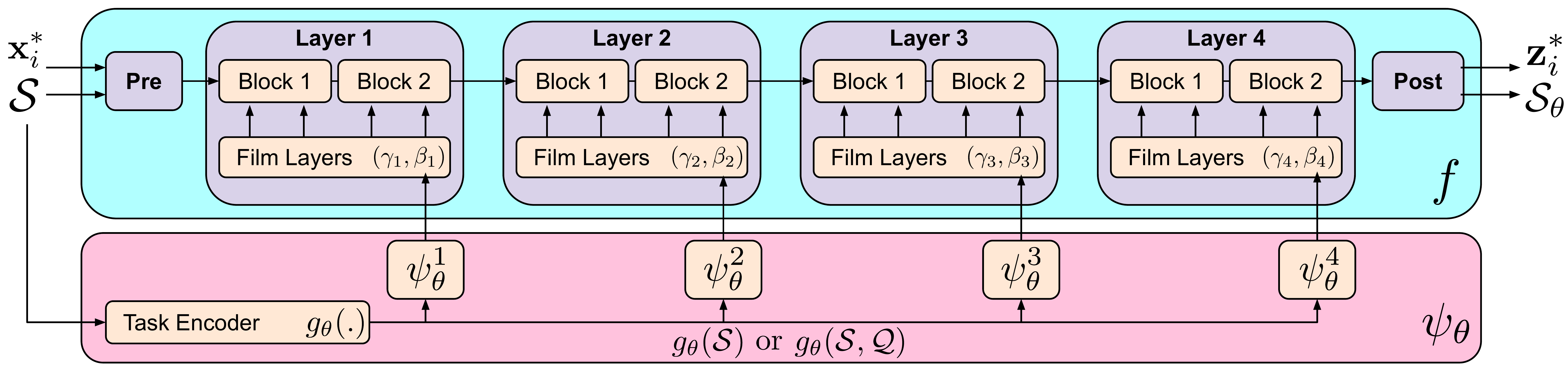}
    \vspace{-0.1in}
    \caption{Overview of neural adaptive feature extraction in Transductive and Simple CNAPS. Figure adapted from \cite{bateni2019improved}.}
    \label{fig:feature-extraction-procedure-overview}
    \vspace{-0.2in}
\end{figure*}

Motivated by these observations, we explore the use of unlabelled examples through transductive learning within the same framework as Simple CNAPS. Our contributions are as follows. \textbf{(1)} We propose a transductive few-shot learner, namely Transductive CNAPS, that extends Simple CNAPS with a transductive two-step task encoder, as well as an iterative soft k-means procedure for refining class parameter estimates (mean and covariance) using both labelled and unlabelled examples. \textbf{(2)} We demonstrate the efficacy of our approach by achieving new state of the art performance on Meta-Dataset \cite{triantafillou2019meta}. \textbf{(3)} When deployed with a feature extractor trained on their respective training sets, Transductive CNAPS achieves state of the art performance on 4 out of 8 settings on mini-ImageNet \cite{Snell17_Proto} and tiered-Imagenet \cite{DBLP:journals/corr/abs-1803-00676-tieredimagenet}, while matching state of the art on another 2. \textbf{(4)} When additional non-overlapping classes from ImageNet \cite{russakovsky2015imagenet} are used to train the feature extractor, Transductive CNAPS is able to leverage this example-rich feature extractor to achieve state of the art across the board on mini-ImageNet and tiered-ImageNet.

%that when deployed at test time, an empirical improvement of 2\%-3\% in classification accuracy is achieved by Transductive CNAPS on the the in-domain tasks on Meta-Dataset \cite{triantafillou2019meta}.
%Furthermore, Transductive CNAPS outperforms CNAPS-based baselines such as Simple CNAPS \cite{bateni2019improved} and CNAPS \cite{requeima2019fast} by statistically significant margins on mini-ImageNet \cite{snell2017prototypical} and tiered-Imagenet \cite{DBLP:journals/corr/abs-1803-00676-tieredimagenet}. 

\section{Related Work}

\subsection{Few-Shot Learning using Labelled Data}
\label{related-work:few-shot-learning}
Early work on few-shot visual classification has focused on improving classification accuracy through the use of better classification metrics with a meta-learned non-adaptive feature extractor. Matching networks \cite{vinyals2016matching} use cosine similarities over feature vectors produced by independently learned feature extractors. Siamese networks \cite{koch2015siamese} classify query images based on the nearest support example in feature space, under the $L_1$ metric. Relation networks \cite{sung2018learning} and variants \cite{DBLP:journals/corr/abs-1905-01436-edge-labelling-gnn, garcia2018fewshot} learn their own similarity metric, parameterised through a Multi-Layer Perceptron. More recently, Prototypical Networks \cite{Snell17_Proto} learn a shared feature extractor that is used to produce class means in a feature space where the Euclidean distance is used for classification. ReMP\footnote{Note that we do not directly compare to these methods as they are either unpublished (ArXiv) or were developed concurrent to our work.} \cite{Zhao2021_ReMP} extends this framework by incorporating self-attention for learning of prototypes in a rectified metric space, maintaining metric consistency between training and testing tasks.

Other work has focused on adapting the feature extractor for new few-shot tasks. Transfer learning by fine-tuning pretrained visual classifiers \cite{DBLP:journals/corr/YosinskiCBL14-finetune} was an early approach that proved limited in success due to issues arising from over-fitting. MAML \cite{finn2017model} and its variants \cite{DBLP:journals/corr/MishraRCA17-snail, DBLP:journals/corr/abs-1803-02999-reptile,DBLP:conf/iclr/RaviL17-meta-lstm} learn meta-parameters that allow fast task-adaptation with only a few gradient updates. Work has also been done on partial adaptation of feature extractors using conditional neural adaptive processes \cite{NIPS2018_7352-tadam, DBLP:journals/corr/abs-1807-01613-cnp, requeima2019fast, bateni2019improved}. These methods rely on channel-wise adaptation of pretrained convolutional layers by adjusting parameters of FiLM layers \cite{perez2018film} inserted throughout the network. Our work builds on the most recent of these neural adaptive approaches, specifically Simple CNAPS \cite{bateni2019improved}. SUR \cite{dvornik2020selecting-sur} and URT \cite{liu2020universal-urt} are two very recent methods that employ universal representations stemming from multiple domain-specific feature extraction heads. URT \cite{liu2020universal-urt}, which was developed and released publicly in parallel to this work, achieves state of the art performance by using a universal transformation layer.

\subsection{Few-Shot Learning using Unlabelled Data}
Several approaches \cite{DBLP:journals/corr/abs-1905-01436-edge-labelling-gnn, DBLP:journals/corr/abs-1805-10002-tpn, DBLP:journals/corr/abs-1803-00676-tieredimagenet} have also explored the use of unlabelled instances for few-shot visual classification. EGNN \cite{DBLP:journals/corr/abs-1905-01436-edge-labelling-gnn} employs a graph convolutional edge-labelling network for iterative propagation of labels from support to query instances. Similarly, TPN \cite{DBLP:journals/corr/abs-1805-10002-tpn} learns a graph construction module for neural propagation of soft labels between elements of the query set.
These methods rely on a neural parameterization of distance within the feature space. TEAM \cite{Qiao_2019_ICCV-team} uses an episodic-wise transductive adaptable metric for performing inference on query examples using a task-specific metric. Song et al. \cite{Song_2020_CVPR-cross-attention} use a cross attention network with a transductive iterative approach for augmenting the support set using the query examples. TAFSSL$^{2}$ \cite{Lichtenstein2020_TAFSSL} improves few-shot learning accuracy in transductive and semi-supervised settings by performing a search for a compact feature sub-space that is discriminative for a given few-shot test-task.

The closest approach to our work is that of Ren et al. \cite{DBLP:journals/corr/abs-1803-00676-tieredimagenet}. Their method extends prototypical networks \cite{Snell17_Proto} by performing a single additional soft-label weighted estimation of class prototypes. Our work, on the other hand, differs in three ways. First, we produce soft-labelled estimates of both class mean and covariance. Second, we use an expectation-maximization (EM) algorithm that performs a dynamic number of soft-label updates, depending on the task at hand. Lastly, we employ a neural-adaptive procedure for feature extraction that is conditioned on a two-step learned transductive task representation, as opposed to a fixed feature-extractor. As we discuss in Section \ref{exp:feot-vs-cot}, this novel task-representation encoder is responsible for substantial performance gains on out-of-domain tasks.

\section{Method}

%\begin{wrapfigure}{r}{3.0in}
%    \centering
%    \includegraphics[width=3.0in]{figures/transductive_set_encoder.pdf}
%    \vspace{-0.1in}
%    \caption{Overview of the transductive task-encoding procedure, $g_\theta(\mathcal{S}, %\mathcal{Q})$, used in Transductive CNAPS.}
%    \label{fig:transductive-set-encoder}
%    \vspace{-0.1in}
%\end{wrapfigure}

\begin{figure}
    \centering
    \includegraphics[width=2.65in]{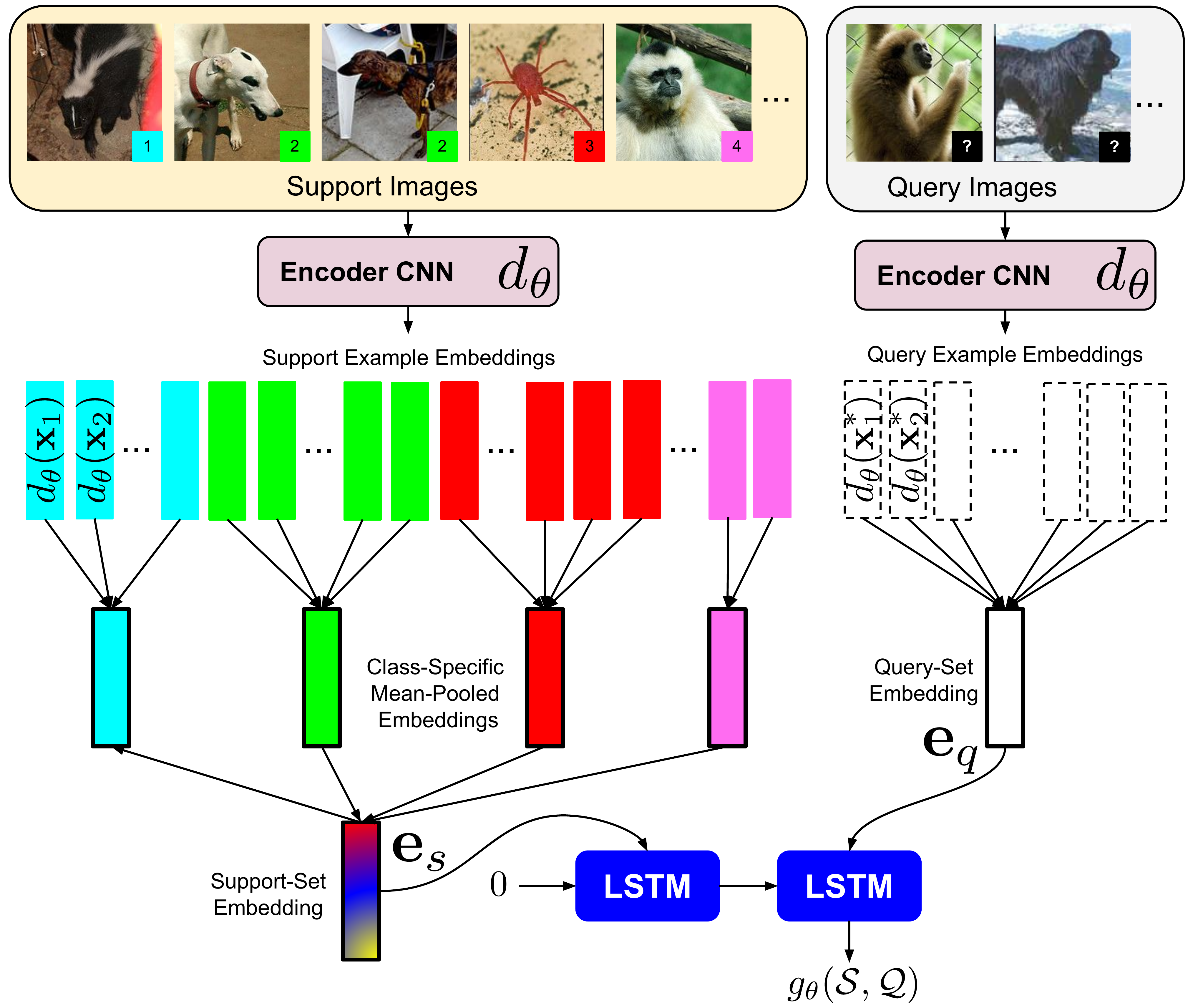}
    \vspace{-0.1in}
    \caption{Overview of the transductive task-encoding procedure, $g_\theta(\mathcal{S}, \mathcal{Q})$, used in Transductive CNAPS.}
    \vspace{-0.2in}
    \label{fig:transductive-task-encoder}
\end{figure}

\subsection{Problem Definition}

Following \cite{Snell17_Proto, bateni2019improved, requeima2019fast, finn2017model}, we focus on a few-shot classification setting where a distribution $D$ over image classification tasks $(\mathcal{S}, \mathcal{Q})$ is provided for training. Each task $(\mathcal{S}, \mathcal{Q}) \sim D$ consists of a support set $\mathcal S = \{(\mathbf{x}_i, y_i)\}_{i=1}^n$ of labelled images and a query set $\mathcal Q = \{\mathbf{x}_i^*\}_{i=1}^{m}$ of unlabelled images; the goal is to predict labels for these query examples, given the (typically small) support set. Each query image $\mathbf{x}_i^* \in \mathcal Q$ has a corresponding ground truth label $y_i^*$ available at training time. A model will be trained by minimizing, over some parameters $\theta$ (which are shared across tasks), the expected query set classification loss over tasks: $\mathbb{E}_{(\mathcal{S}, \mathcal{Q}) \sim D}[\sum_{\mathbf{x}_i^* \in \mathcal Q}  - \log p_\theta(y^*_i|\mathbf x^*_i, \mathcal{S}, \mathcal{Q})]$; the inclusion of the dependence on all of $\mathcal{Q}$ here allows for the model to be transductive. At test time, a separate distribution of tasks generated from previously unseen images and classes is used to evaluate performance. Let us also define \textit{shot} as the number of support examples per class, and \textit{way} as the number of classes within the task.

\subsection{Simple CNAPS}
\label{method:simple-cnaps}

Our method extends the Simple CNAPS \cite{bateni2019improved} architecture for few-shot visual classification. Simple CNAPS performs few-shot classification in two steps. 

First, it computes task-adapted features for every support and query example. This part of the architecture is the same as that in CNAPS \cite{requeima2019fast}, and is based on the FiLM meta-learning framework \cite{perez2018film}. Second, it uses the support set to estimate a per-class Mahalanobis metric, which is used to assign query examples to classes. The architecture uses a ResNet18 \cite{DBLP:journals/corr/HeZRS15-resnet} feature extractor. Within each residual block, Feature-wise Linear Modulation (FiLM) layers compute a scale factor $\gamma$ and shift $\beta$ for each output channel, using block-specific adaptation networks $\psi_\theta$ that are conditioned on a task encoding. The task encoding $g_\theta(\mathcal{S})$ consists of the mean-pooled feature vectors of support examples produced by $d_\theta$, a separate but end-to-end learned Convolution Neural Network (CNN). This produces an adapted feature extractor $f_\theta$ (which implicitly depends on the support set $\mathcal{S}$) that maps support/query images onto the corresponding adapted feature space.  We will denote by $\mathcal{S}_\theta, \mathcal{Q}_\theta$ versions of the support/query sets where each image is mapped into its feature representation $\mathbf z = f_\theta(\mathbf x)$. 

%\begin{figure*}
%    \centering
%    \subfloat[Transductive Set-Encoder Overview]{{\includegraphics[width=2.45in]{figures/transductive_set_encoder.pdf} }}
%    \subfloat[Transductive CNAPS vs. Simple CNAPS]{{\includegraphics[width=2.95in]{figures/simple-cnaps-vs-transductive-cnaps.pdf} }}
%    \vspace{-0.1in}
%    \caption{a) Overview of the transductive task-encoding procedure, $g_\theta(\mathcal{S}, \mathcal{Q})$, used in Transductive CNAPS. b) Transductive CNAPS (right) extends the Mahalanobis-distance based classifier in Simple CNAPS (left) through transductive soft k-meanshttps://www.overleaf.com/project/600f5a327b9acb2ab863dbf6 clustering of the visual space.}
%    \label{fig:transductive-cnaps-vs-simple-cnaps}
%    \vspace{-0.2in}
%\end{figure*}

\begin{figure*}
    \centering
    \includegraphics[width=0.735\textwidth]{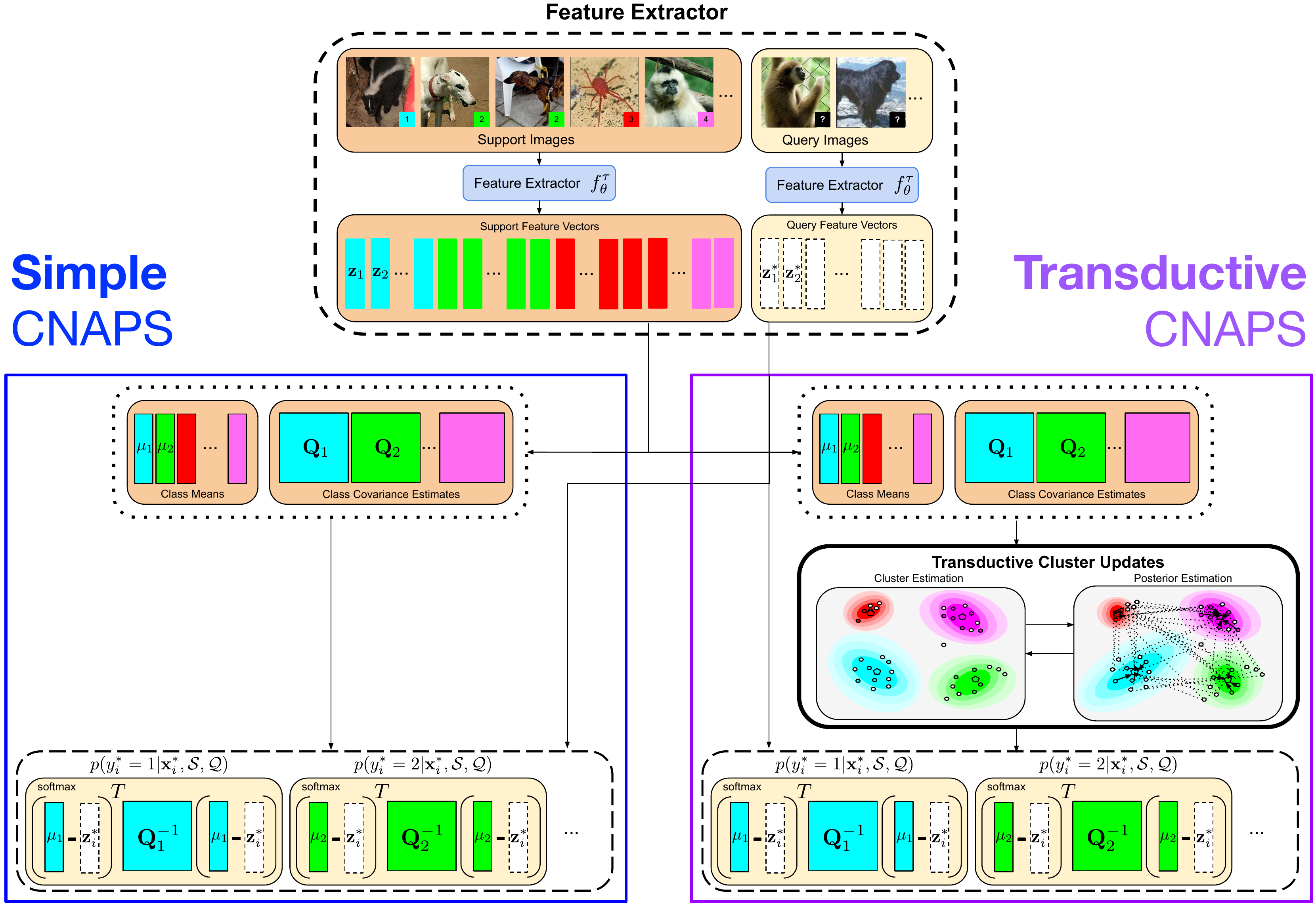}
    \vspace{-0.1in}
    \caption{Transductive CNAPS (right) extends the Mahalanobis-distance based classifier in Simple CNAPS (left) through transductive soft k-means clustering of the visual space.}
    \vspace{-0.2in}
    \label{fig:transductive-cnaps-vs-simple-cnaps}
\end{figure*}

\begin{figure*}%
    \vspace{-0.1in}
    \centering
    \subfloat[In-Domain]{{\includegraphics[width=1.95in]{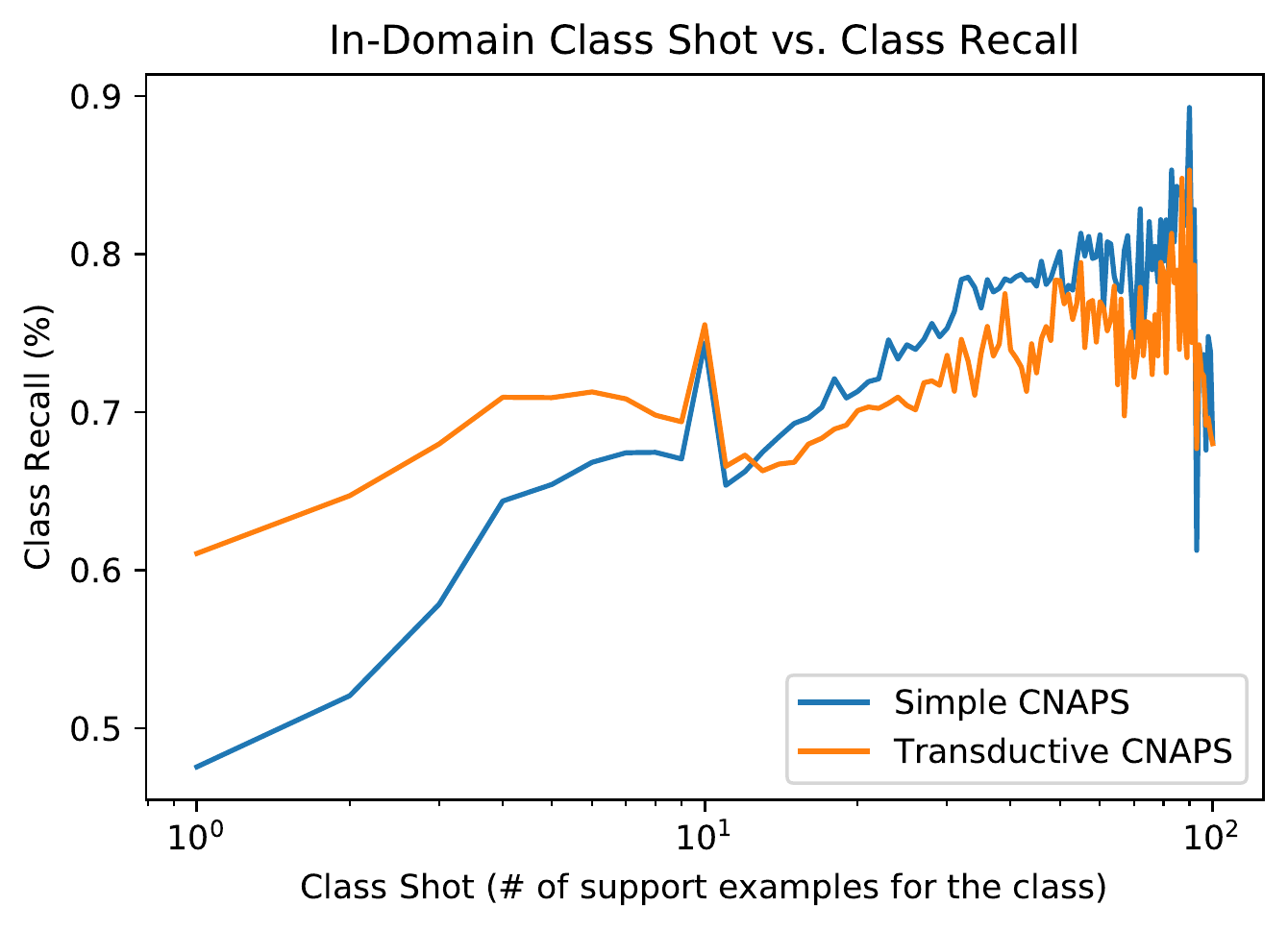} }}
    \subfloat[Out-of-Domain]{{\includegraphics[width=1.95in]{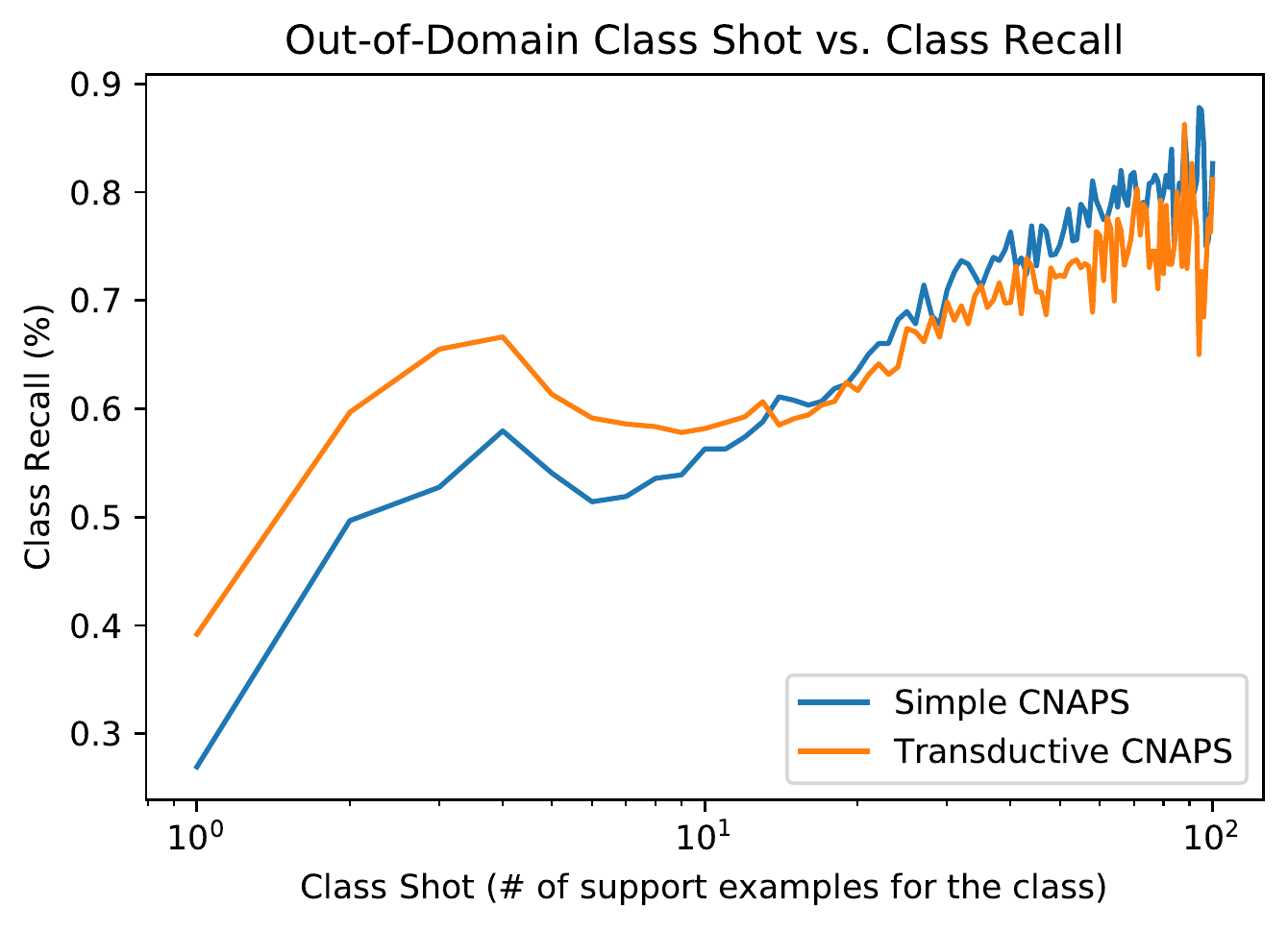} }}
    \subfloat[Overall]{{\includegraphics[width=1.95in]{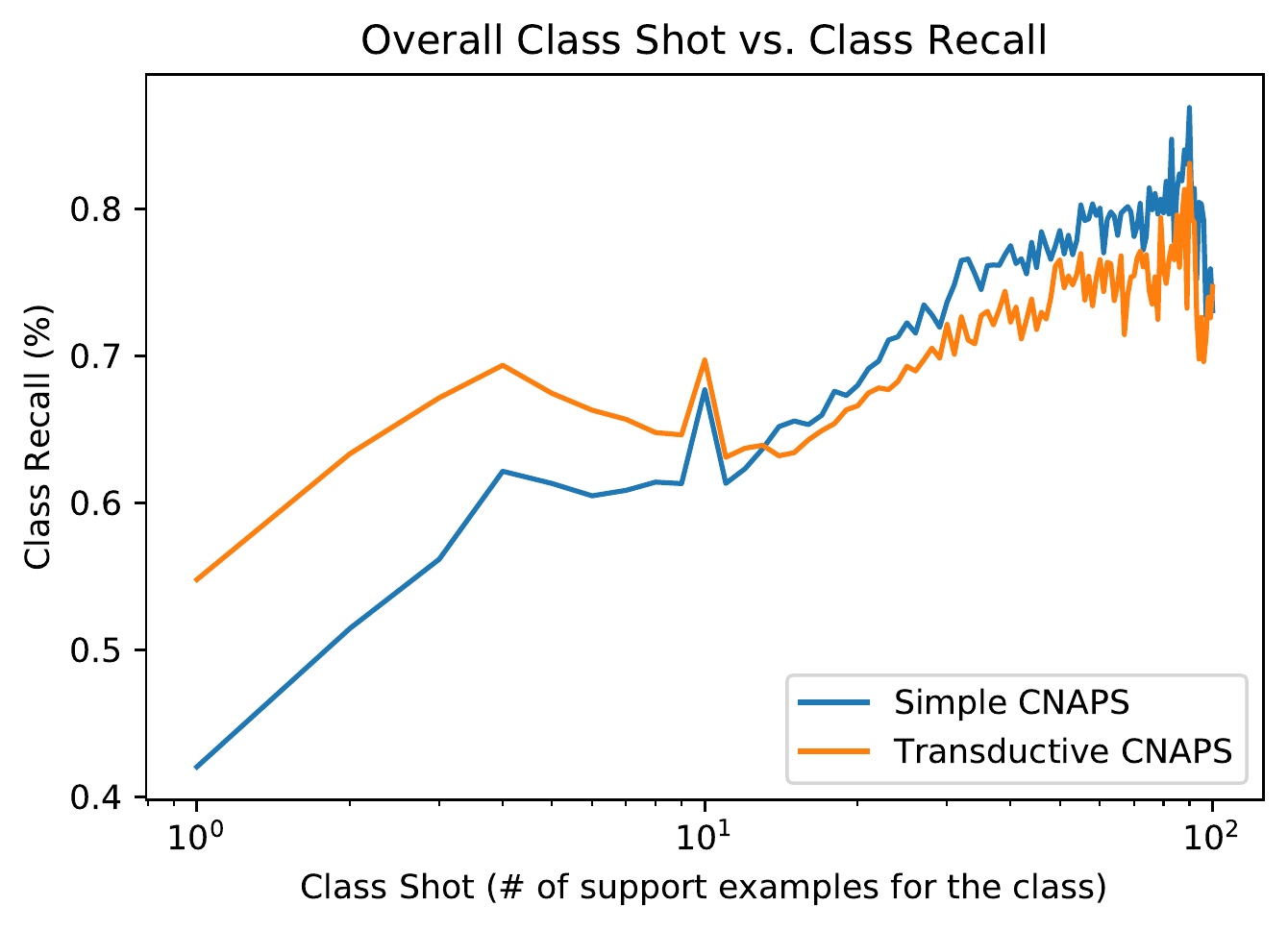} }}\\%
    \vspace{-0.12in}
    \caption{Class recall (otherwise noted as in-class query accuracy) averaged between classes across all tasks and (a: In-Domain, 
    b: Out-of-domain, c: all) Meta-Dataset sub-datasets. Class recalls have been grouped together, averaged and plotted according to class shot in (a), (b), and (c).}
    \vspace{-0.2in}
    \label{fig:ratios-and-shots-all-ranges-10max}
\end{figure*}

Simple CNAPS then computes a Mahalanobis distance relative to each class $k$ by estimating a mean $\bm{\mu}_k$ and regularized covariance $\mathbf{Q}_k$ in the adapted feature space, using the support instances:
\begin{align}
    \bm{\mu}_k 
    &= 
    \frac{1}
         {n_k} 
    \sum_{i}
    \: 
    \mathbb{I}[y_i=k] \:
    \mathbf z_i
    \\
    \mathbf{Q}_k 
    &=
    \lambda_k \,
    \bm{\Sigma}_k
    +
    (1 - \lambda_k) \,
    \bm{\Sigma}
    +
    \beta I,
    &
    \lambda_k
    &=
    \frac{n_k}
         {n_k + 1}
    \label{simple-cnaps-eq:mean-cov-calculation}
\end{align}
Here $\mathbb{I}[y_i=k]$ is the indicator function and $n_k = \sum_i \mathbb I[y_i=k]$ is the number of examples from class $k$ in the support set $\mathcal{S}$. The ratio $\lambda_k$ balances a task-conditional sample covariance $\bm{\Sigma}$ and a class-conditional sample covariance $\bm{\Sigma}_k$:
\begin{align}
    \bm{\Sigma}
    &= 
    \frac{1}{n}
    \sum_{i}
    \!
    \big( \mathbf z_i \!-\! \bm{\mu}\big)
    \big( \mathbf z_i \!-\! \bm{\mu} \big)^T
    \\
    \bm{\Sigma}_k
    &= 
    \frac{1}{n_k}
    \sum_{i}
    \mathbb{I}[y_i=k]\:
    \big( \mathbf z_i \!-\! \bm{\mu}_k \big)
    \big( \mathbf z_i \!-\! \bm{\mu}_k \big)^T
\end{align}
where $\bm{\mu} = \frac{1}{n}\sum_{i} \mathbf z_i$ is the task-level mean. When few support examples are available for a particular class, $\lambda_k$ is small, and the estimate is regularized towards the task-level covariance $\bm{\Sigma}$. As the number of support examples for the class increases, the estimate tends towards the class-conditional covariance $\bm{\Sigma}_k$. Additionally, a regularizer $\beta I$ (we set $\beta=1$ in our experiments) is added to ensure invertibility. Given the class means and covariances, Simple CNAPS computes class probabilities for each query feature vector $\mathbf{z}^*_i$ through a softmax over the squared Mahalanobis distances with respect to each class:
\begin{align}
    p(y^* = k \mid \mathbf z^*) \propto \exp
    \big(
      -(\mathbf z - \bm{\mu}_k)^T\mathbf{Q}_k^{-1}(\mathbf z - \bm{\mu}_k)
    \big)
    \label{simple-cnaps-eq:simple-cnaps-mahalanobis-classifier}
\end{align}

\subsection{Transductive CNAPS}

Transductive CNAPS extends Simple CNAPS by taking advantage of the query set, both in the feature adaptation step and the classification step. First, the task encoder $g_\theta$ is extended to incorporate both a support-set embedding $\mathbf{e}_s$ and a query-set embedding $\mathbf{e}_q$ such that,
\begin{align}
    \mathbf{e}_s 
    &= 
    \frac{1}{K} \sum_{k} \frac{1}{n_k} \sum_{i}
    \mathbb{I}[y_i=k]\: d_\theta(\mathbf{x}_i),
    \\
    \mathbf{e}_q 
    &= 
    \frac{1}{n_q} \sum_{i*} d_\theta(\mathbf{x}_i^*),
    \label{transductive-cnaps-eq:encoder}
\end{align}
where $d_\theta$ is a learned CNN. The support embedding $\mathbf{e}_s$ is formed by an average of (encoded) support examples, with weighting inversely proportional to their class counts to prevent bias from class imbalance. The query embedding $\mathbf e_q$ uses simple mean-pooling; both $\mathbf{e}_s$ and $\mathbf{e}_q$ are invariant to permutations of the respective support/query instances. We then process $\mathbf e_s$ and $\mathbf e_q$ through two steps of a Long Short Term Memory (LSTM) network in the same order to generate the final transductive task-embedding $g_\theta(\mathcal{S}, \mathcal{Q})$ used for adaptation. This process is visualized in Figure \ref{fig:transductive-task-encoder}.

Second, we can interpret Simple CNAPS as a form of ``supervised clustering'' in feature space; each cluster (corresponding to a class $k$) is parameterized with a centroid $\bm{\mu}_k$ and a metric $\mathbf Q_k^{-1}$, and we interpret \eqref{simple-cnaps-eq:simple-cnaps-mahalanobis-classifier} as class assignment probabilities based on the distance to each centroid. With this viewpoint in mind, a natural extension to consider is to use the estimates of the class assignment probabilities on unlabelled data to refine the class parameters $\bm{\mu}_k, \mathbf Q_k$ in a soft $k$-means framework based on per-cluster Mahalanobis distances \cite{melnykov2014k}.  In this framework, as shown in Figure \ref{fig:clustering-method}, we alternate between computing updated assignment probabilities using \eqref{simple-cnaps-eq:simple-cnaps-mahalanobis-classifier} on the query set and using those assignment probabilities to compute updated class parameters.

We define $\mathcal{R}_\theta = \mathcal{S}_\theta \sqcup \mathcal{Q}_\theta$ as the disjoint union of the support set and the query set. For each element of $\mathcal{R}_\theta$, which we index by $j$, we define responsibilities $w_{jk}$ in terms of their class predictions when it is part of the query set and in terms of the label when it is part of the support set,
\begin{align}
    w_{jk} = \begin{cases}
        p\big(y_j'=k \mid \mathbf z_j'\big)
        & \mathbf{z}_j'\in \mathcal{Q}_\theta,
        \\
        \mathbb{I}[y'_j = k]
        &
        (\mathbf{z}_j', y_j') \in \mathcal{S}_\theta.
       \end{cases} 
    \label{eq:class_weights}
\end{align}
Using these responsibilities we can incorporate unlabelled samples from the support set by defining weighted estimates $\bm{\mu}'_k$ and $\mathbf{Q}'_k$:
\begin{align}
    \label{eq:update_first}
    \bm{\mu}'_k 
    &= 
    \frac{1}{n'_k}
    \sum_{j} 
    w_{jk} \:
    \mathbf z_j',
    \\
    \mathbf{Q}'_k 
    &= 
    \lambda'_k 
    \bm{\Sigma}'_k 
    + 
    (1 - \lambda'_k) 
    \bm{\Sigma}' 
    + \beta I
    ,
\end{align}
where $n'_k = \sum_{j} w_{jk}$ defines $\lambda'_k = n'_k / (n'_k + 1)$. The covariance estimates $\bm{\Sigma}'$ and $\bm{\Sigma}_k'$ are 

\begin{align}
    \label{eq:update_last}
    \bm{\Sigma}'
    &= 
    \frac{1}{\sum_k n'_k}
    \sum_{jk}
    \!
    w_{jk}
    \big( \mathbf z_j' \!-\! \bm{\mu}' \big)
    \big( \mathbf z_j' \!-\! \bm{\mu}' \big)^T
    \\
    \bm{\Sigma}'_k
    &= 
    \frac{1}{n'_k}
    \sum_{j}
    \!
    w_{jk}
    \big( \mathbf z_j' \!-\! \bm{\mu}'_k \big)
    \big( \mathbf z_j' \!-\! \bm{\mu}'_k \big)^T
\end{align}
where $\bm{\mu}' = \left(\sum_{k} n_k'\right)^{-1}\sum_{jk}w_{jk}\mathbf z_j'$ is the task mean.

\begin{algorithm*}[t]
    \caption{Iterative Refinement in Transductive CNAPS}
    \label{algo:soft-kmeans-cnaps}
    \begin{algorithmic}[1] % The number tells where the line numbering should start
        \Procedure{compute\_query\_labels}{$\mathcal{S}_\theta,\mathcal{Q}_\theta, N_\text{iter}$}
            \State For $j$ ranging over support and query sets, $w_{jk} \gets \begin{cases}
                1 &\text{if}~ (\mathbf{z}'_j,y'_j) \in \mathcal{S}_\theta \text{~and~} y_j=k 
                \\ 0 &\text{otherwise}
            \end{cases}$ 
            \For{iter = $0 \cdots N_\text{iter}$}\Comment The first iteration is equivalent to Simple CNAPS;
            % \While{$\neg$converged}
                % \State \Comment Further iterations refine the class estimates.
                \State Compute class parameters $\bm{\mu}_k, \mathbf Q_k$ according to update equations \eqref{eq:update_first}-\eqref{eq:update_last}
                \State Compute class weights using class parameters according to \eqref{eq:class_weights}
                \State \textbf{break} if the most probable class for each query example hasn't changed
                % \begin{enumerate}
                %     \item We have hit a maximum iteration count, or
                %     \item We have iterated at least twice and the most probable class assignments have not changed since the last iteration.
                % \end{enumerate}
            % \EndWhile
            \EndFor
            \State \Return class probabilities $w_{jk}$ for $j$ corresponding to $\mathcal{Q}_\theta$
        \EndProcedure
    \end{algorithmic}
    \label{alg:iterative}
\end{algorithm*}

These update equations are weighted versions of the original Simple CNAPS estimators from Section \ref{method:simple-cnaps}, and reduce to them exactly in the case of an empty query set. 

Algorithm \ref{alg:iterative} summarizes the soft k-means procedure based on these updates. We initialize our weights using only the labelled support set. We use those weights to compute class parameters, then compute updated weights using both the support and query sets. At this point, the weights associated with the query set $\mathcal{Q}$ are the same class probabilities as estimated by Simple CNAPS.  However, we continue this procedure iteratively until we reach either reach a maximum number of iterations, or until class assignments $\text{argmax}_k \, w_{jk}$ stop changing. 

\begin{figure}
  \vspace{-0.075in}
  \begin{center}
    \includegraphics[width=0.4\textwidth]{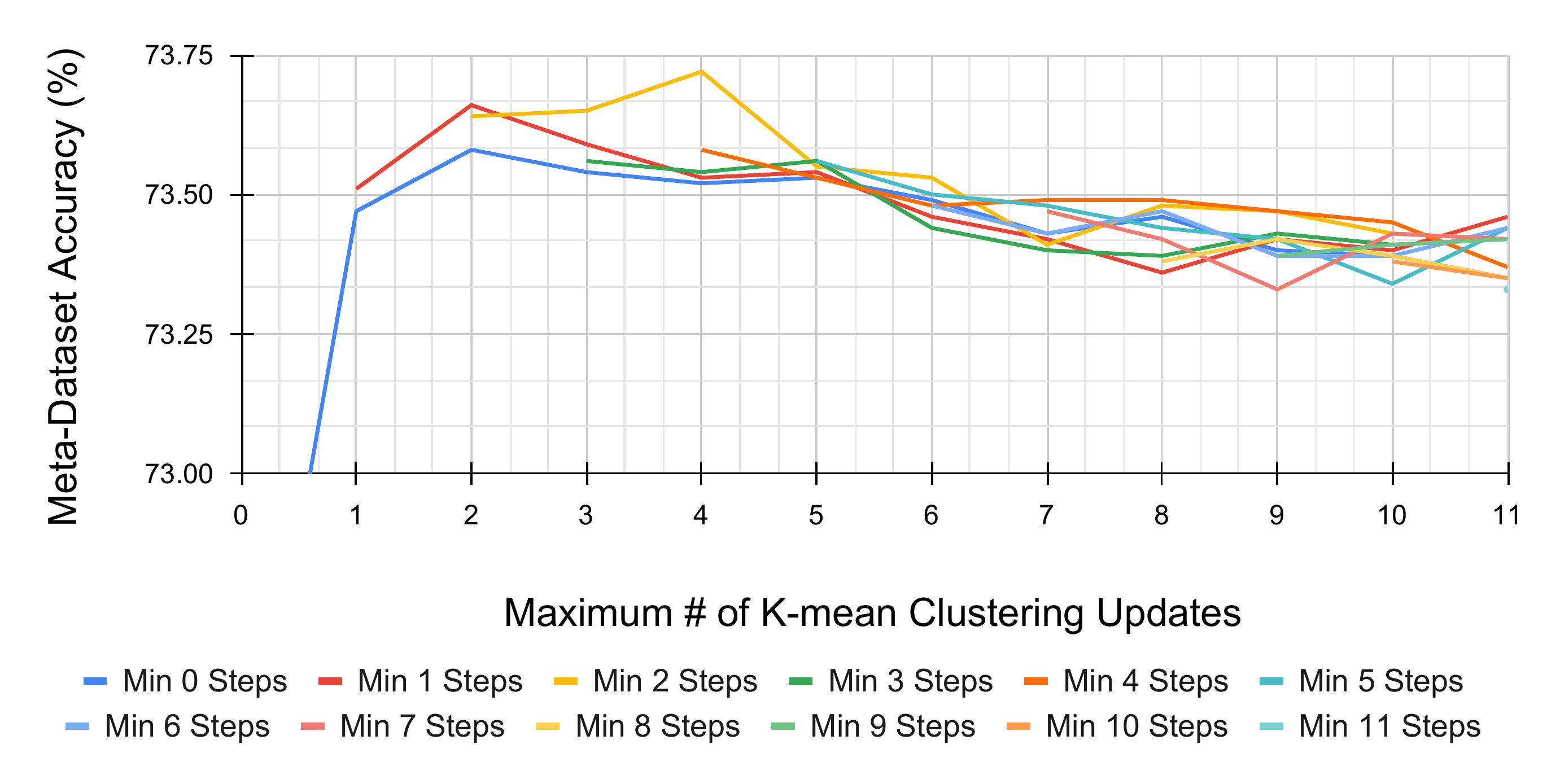}
  \end{center}
  \vspace{-0.3in}
  \caption{Evaluating Transductive CNAPS on Meta-Dataset with different minimum and maximum number of steps. Performances reported stem from five run averages.}
  \label{fig:max-and-min}
  \vspace{-0.3in}
\end{figure}

Unlike the transductive task-encoder, this second extension, namely the soft k-mean iterative estimation of class parameters, is used at test time only. During training, a single estimation is produced for both mean and covariance using only the support examples. This, as we discuss more in Section \ref{exp:feot-vs-cot}, was shown to empirically perform better. See Figure \ref{fig:transductive-cnaps-vs-simple-cnaps} for a high-level visual comparison of classification in Simple CNAPS vs. Transductive CNAPS.

\subsection{Relationship to Bregman Soft Clustering}

The procedure in Algorithm~\ref{algo:soft-kmeans-cnaps} resembles the Bregman clustering algorithms proposed by Banerjee et al.~\cite{banerjee2005clustering}. Specifically, the updates to soft assignments $w_{jk}$ in Equation \ref{simple-cnaps-eq:simple-cnaps-mahalanobis-classifier} are the semi-supervised equivalent of those in Bregman soft clustering, in which the divergence is based on the Mahalanobis distance  $F(\textbf{z}) = \textbf{z}^T \textbf{Q}^{-1} \textbf{z}$,
\begin{align}
    D_F(\textbf{z}, \textbf{z}') = F(\textbf{z}) - F(\textbf{z}') - \nabla F(\textbf{z}')^T(\textbf{z} - \textbf{z}').
\end{align}
However, Algorithm~\ref{algo:soft-kmeans-cnaps} differs in that it updates both $\bm{\mu}'_k$ and $\textbf{Q}'_k$ at each iteration, rather than just $\bm{\mu}'_k$.

In general, any (regular) exponential family can be associated with a Bregman divergence and vice versa, which gives rise to a correspondence between EM-based clustering and Bregman soft clustering algorithms \cite{banerjee2005clustering}. Standard Bregman soft clustering corresponds to EM in which the likelihood is a Gaussian with unknown mean and a known covariance $\textbf{Q}$ that is shared across clusters. The case where the covariance is unknown corresponds to Gaussian mixture models (GMMs), but the function $F(\textbf{z})$ is not simply the Mahalanobis distance in this case.

The updates for $\bm{\mu}'_k$ and $\textbf{Q}'_k$ in Algorithm~\ref{algo:soft-kmeans-cnaps} are equivalent to those in a GMM that incorporates regularization for the covariances. However, GMM clustering differs in the calculation of the assignment probabilities 
\begin{align*}
    p(y^* &= k \mid \mathbf z^*) \propto \pi_k\\ &\exp
    \left(
      -\frac{1}{2}(\mathbf z - \bm{\mu_k})^T\mathbf{Q}_k^{-1}(\mathbf z - \bm{\mu}_k)-\frac{1}{2}\log |\mathbf Q_k|\right).
\end{align*}
These probabilities incorporate a term $\pi_k = p(y^* = k)$, which defines a prior probability of assignments to a cluster, and a term $\exp(- \log |\textbf{Q}_k|)$, which reflects the fact that GMMs employ a likelihood with unknown covariance.

In short, our clustering procedure employs an update to soft assignments $w_{jk}$ that is similar to the one in soft Bregman clustering, but employs updates to $\bm{\mu}'_{k}$ and $\textbf{Q}'_k$ that are similar to those in a (regularized) GMM. In Section \ref{exp:gaussian-mixture-models} we demonstrate through ablations that this combination of updates improves emperical performance relative to baselines that perform GMM-based clustering.

\section{Experiments}

\subsection{Benchmarks}
\label{experiments:benchmarks}

\noindent
\textbf{Meta-Dataset} \cite{triantafillou2019meta} is a few-shot visual classification benchmark that consists of 10 widely used datasets: ILSVRC-2012 (ImageNet) \cite{russakovsky2015imagenet}, Omniglot \cite{lake2015human}, FGVC-Aircraft (Aircraft) \cite{maji2013fine}, CUB-200-2011 (Birds) \cite{wah2011caltech}, Describable Textures (DTD) \cite{cimpoi2014describing}, QuickDraw \cite{jongejan2016quick}, FGVCx Fungi (Fungi)  \cite{fungi2018schroeder}, VGG Flower (Flower) \cite{nilsback2008automated}, Traffic Signs (Signs) \cite{houben2013detection} and MSCOCO \cite{lin2014microsoft}. Consistent with past work \cite{requeima2019fast, bateni2019improved}, we train our model on the official training splits of the first 8 datasets and use the test splits to evaluate in-domain performance. We use the remaining two datasets as well as three external benchmarks, namely MNIST \cite{lecun-mnisthandwrittendigit-2010}, CIFAR10 \cite{Krizhevsky09learningmultiple} and CIFAR100 \cite{Krizhevsky09learningmultiple}, for out-of-domain evaluation. 

Task generation in Meta-Dataset follows a complex procedure where tasks can be of different \textit{ways} and individual classes can be of varying \textit{shots} even within the same task. Specifically, for each task, the task \textit{way} is first sampled uniformly between 5 and 50 and \textit{way} classes are selected at random from the corresponding class/dataset split. Then, for each class, 10 instances are sampled at random and used as query examples for the class, while of the remaining images for the class, a \textit{shot} is sampled uniformly from [1, 100] and \textit{shot} number of images are selected at random as support examples with total support set size of 500. 

Additional dataset-specific constraints are enforced, as discussed in Section 3.2 of \cite{triantafillou2019meta}, and since some datasets have fewer than 50 classes and fewer than 100 images per class, the overall \textit{way} and \textit{shot} distributions resemble Poisson distributions where most tasks have fewer than 10 classes and most classes have fewer than 10 support examples (see Appendix-\ref{appendix:benchmark-and-training:meta-dataset}). Following \cite{bateni2019improved} and \cite{requeima2019fast}, we first train our ResNet18 feature extractor on the Meta-Dataset defined training split of ImageNet following the procedure in Appendix-\ref{appendix:benchmark-and-training:meta-dataset-training-and-testing}. The ResNet18 parameters are then kept fixed while we train the adaptation network for a total of 110K sampled tasks using Episodic Training \cite{Snell17_Proto, finn2017model} (see Appendix-\ref{appendix:benchmark-and-training:meta-dataset-training-and-testing} for details).

\vspace{0.02in}
\noindent
\textbf{Mini/tiered-ImageNet} \cite{vinyals2016matching, DBLP:journals/corr/abs-1803-00676-tieredimagenet} are two benchmarks for few-shot learning. Both datasets employ subsets of ImageNet \cite{russakovsky2015imagenet} with a total of 100 classes and 60K images in mini-ImageNet and 608 classes and 779K images in tiered-ImageNet. Unlike Meta-Dataset, tasks across these datasets have pre-defined \textit{shots} and \textit{ways} that are uniform across every task generated in the specified setting. 

Following \cite{DBLP:journals/corr/abs-1803-02999-reptile, DBLP:journals/corr/abs-1805-10002-tpn, Snell17_Proto}, we report performance on the 1/5-\textit{shot} 5/10-\textit{way} settings across both datasets with 10 query examples per class. We first train the ResNet18 on the training set of the corresponding benchmark at hand following the procedure noted in Appendix-\ref{appendix:benchmark-and-training:minitiered-training-and-testing}. We also consider a more feature-rich ResNet18 trained on the larger ImageNet dataset. However, we exclude classes and examples from test sets of mini/tiered-ImageNet to address potential class/example overlap issues, resulting in 825 classes and 1,055,494 images remaining. Then, with the ResNet18 parameters fixed, we train episodically for 20K tasks (see Appendix-\ref{appendix:benchmark-and-training:minitiered} for details).

\subsection{Results}

\begin{table*}[t]
    \centering
    \scriptsize
    \tabcolsep=0.09cm
    \begin{tabular}{lcccccccccccccccc}
        & \multicolumn{8}{c}{In-Domain Accuracy (\%)} & \multicolumn{5}{c}{Out-of-Domain Accuracy (\%)} & \multicolumn{3}{c}{Avg Rank} \\
        \cmidrule(lr){2-9}\cmidrule(lr){10-14}
        Model & \rotatebox{45}{ImageNet} & \rotatebox{45}{Omniglot} & \rotatebox{45}{Aircraft} & \rotatebox{45}{Birds} & \rotatebox{45}{DTD} & \rotatebox{45}{QuickDraw} & \rotatebox{45}{Fungi} & \rotatebox{45}{Flower} & \rotatebox{45}{Signs} & \rotatebox{45}{MSCOCO} & \rotatebox{45}{MNIST} & \rotatebox{45}{CIFAR10} & \rotatebox{45}{CIFAR100} & In & Out & All \\
        \midrule
        RelationNet & 30.9\textpm0.9 & 86.6\textpm0.8 & 69.7\textpm0.8 & 54.1\textpm1.0 & 56.6\textpm0.7 & 61.8\textpm1.0 & 32.6\textpm1.1 & 76.1\textpm0.8 & 37.5\textpm0.9 & 27.4\textpm0.9 & NA & NA & NA & 10.5 & 11.0 & 10.6 \\
        MatchingNet & 36.1\textpm1.0 & 78.3\textpm1.0 & 69.2\textpm1.0 & 56.4\textpm1.0 & 61.8\textpm0.7 & 60.8\textpm1.0 & 33.7\textpm1.0 & 81.9\textpm0.7 & 55.6\textpm1.1 & 28.8\textpm1.0 & NA & NA & NA & 10.1 & 8.5 & 9.8 \\
        MAML & 37.8\textpm1.0 & 83.9\textpm1.0 & 76.4\textpm0.7 & 62.4\textpm1.1 & 64.1\textpm0.8 & 59.7\textpm1.1 & 33.5\textpm1.1 & 79.9\textpm0.8 & 42.9\textpm1.3 & 29.4\textpm1.1 & NA & NA & NA & 9.2 & 10.5 & 9.5 \\
        ProtoNet & 44.5\textpm1.1 & 79.6\textpm1.1 & 71.1\textpm0.9 & 67.0\textpm1.0 & 65.2\textpm0.8 & 64.9\textpm0.9 & 40.3\textpm1.1 & 86.9\textpm0.7 & 46.5\textpm1.0 & 39.9\textpm1.1 & NA & NA & NA & 8.2 & 9.5 & 8.5 \\
        ProtoMAML & 46.5\textpm1.1 & 82.7\textpm1.0 & 75.2\textpm0.8 & 69.9\textpm1.0 & 68.3\textpm0.8 & 66.8\textpm0.9 & 42.0\textpm1.2 & 88.7\textpm0.7 & 52.4\textpm1.1 & 41.7\textpm1.1 & NA & NA & NA & 7.1 & 8.0 & 7.3 \\
        CNAPS & 52.3\textpm1.0 & 88.4\textpm0.7 & 80.5\textpm0.6 & 72.2\textpm0.9 & 58.3\textpm0.7 & 72.5\textpm0.8 & 47.4\textpm1.0 & 86.0\textpm0.5 & 60.2\textpm0.9 & 42.6\textpm1.1 & 92.7\textpm0.4 & 61.5\textpm0.7 & 50.1\textpm1.0 & 6.6 & 6.0 & 6.4 \\
        BOHB-E & 55.4\textpm1.1 & 77.5\textpm1.1 & 60.9\textpm0.9 & 73.6\textpm0.8 & 72.8\textpm0.7 & 61.2\textpm0.9 & 44.5\textpm1.1 & 90.6\textpm0.6 & 57.5\textpm1.0 & 51.9\textpm1.0 & NA & NA & NA & 6.4 & 4.0 & 5.9 \\
        TaskNorm & 50.6\textpm1.1 & 90.7\textpm0.6 & 83.8\textpm0.6 & 74.6\textpm0.8 & 62.1\textpm0.7 & 74.8\textpm0.7 & 48.7\textpm1.0 & 89.6\textpm0.5 & 67.0\textpm0.7 & 43.4\textpm1.0 & 92.3\textpm0.4 & 69.3\textpm0.8 & 54.6\textpm1.1 & 4.7 & 4.8 & 4.8 \\
        Simple CNAPS & \textbf{58.6\textpm1.1} & 91.7\textpm0.6 & 82.4\textpm0.7 & 74.9\textpm0.8 & 67.8\textpm0.8 & 77.7\textpm0.7 & 46.9\textpm1.0 & 90.7\textpm0.5 & 73.5\textpm0.7 & 46.2\textpm1.1 & 93.9\textpm0.4 & 74.3\textpm0.7 & 60.5\textpm1.0 & 3.4 & 3.0 & 3.2 \\
        SUR & 56.3\textpm1.1 & 93.1\textpm0.5 & \textbf{85.4\textpm0.7} & 71.4\textpm1.0 & \textbf{71.5\textpm0.8} & 81.3\textpm0.6 & \textbf{63.1\textpm1.0} & 82.8\textpm0.7 & 70.4\textpm0.8 & \textbf{52.4\textpm1.1} & 94.3\textpm0.4 & 66.8\textpm0.9 & 56.6\textpm1.0 & 3.1 & 2.6 & 2.9 \\
        URT & 55.7\textpm1.0 & \textbf{94.4\textpm0.4} & \textbf{85.8\textpm0.6} & \textbf{76.3\textpm0.8} & \textbf{71.8\textpm0.7} & \textbf{82.5\textpm0.6} & \textbf{63.5\textpm1.0} & 88.2\textpm0.6 & 69.4\textpm0.8 & \textbf{52.2\textpm1.1} & 94.8\textpm0.4 & 67.3\textpm0.8 & 56.9\textpm1.0 & \textbf{1.7} & 2.8 & 2.2 \\
        \midrule
        Our Method & \textbf{58.8\textpm1.1} & \textbf{93.9\textpm0.4} & 84.1\textpm0.6 & \textbf{76.8\textpm0.8} & 69.0\textpm0.8 & 78.6\textpm0.7 & 48.8\textpm1.1 & \textbf{91.6\textpm0.4} & \textbf{76.1\textpm0.7} & 48.7\textpm1.0 & \textbf{95.7\textpm0.3} & \textbf{75.7\textpm0.7} & \textbf{62.9\textpm1.0} & 2.1 & \textbf{1.6} & \textbf{1.9} \\
    \end{tabular}
    \vspace{-0.1in}
    \caption{Few-shot classification on Meta-Dataset, MNIST, and CIFAR10/100. Error intervals correspond to 95\% confidence intervals, and bold values indicate statistically significant state of the art performance. Average rank is obtained by ranking methods on each dataset and averaging the ranks.}
    \vspace{-0.2in}
    \label{tab:results:meta-dataset}
\end{table*}{}

\begin{table*}[t]
    \centering
    \scriptsize
    \tabcolsep=0.3cm
    \begin{tabular}{lcccccccccc}
        {} & {} & \multicolumn{4}{c}{mini-ImageNet Accuracy (\%)} & \multicolumn{4}{c}{tiered-ImageNet Accuracy (\%)} \\
        \cmidrule(lr){3-6}\cmidrule(lr){7-10}
        {} & {} & \multicolumn{2}{c}{5-\textit{way}} & \multicolumn{2}{c}{10-\textit{way}} & \multicolumn{2}{c}{5-\textit{way}} & \multicolumn{2}{c}{10-\textit{way}}\\
        \cmidrule(lr){3-4}\cmidrule(lr){5-6}\cmidrule(lr){7-8}\cmidrule(lr){9-10}
        Model & Transductive & 1-\textit{shot} & 5-\textit{shot} & 1-\textit{shot} & 5-\textit{shot} & 1-\textit{shot} & 5-\textit{shot} & 1-\textit{shot} & 5-\textit{shot} \\
        \midrule
        MAML \cite{finn2017model} & BN & 48.7\textpm1.8 & 63.1\textpm0.9 & 31.3\textpm1.1 & 46.9\textpm1.2 & 51.7\textpm1.8 & 70.3\textpm1.7 & 34.4\textpm1.2 & 53.3\textpm1.3 \\
        MAML+ \cite{DBLP:journals/corr/abs-1805-10002-tpn} & Yes & 50.8\textpm1.8 & 66.2\textpm1.8 & 31.8\textpm0.4 & 48.2\textpm1.3 & 53.2\textpm1.8 & 70.8\textpm1.8 & 34.8\textpm1.2 & 54.7\textpm1.3 \\
        Reptile \cite{DBLP:journals/corr/abs-1803-02999-reptile} & No & 47.1\textpm0.3 & 62.7\textpm0.4 & 31.1\textpm0.3 & 44.7\textpm0.3 & 49.0\textpm0.2 & 66.5\textpm0.2 & 33.7\textpm0.3 & 48.0\textpm0.3 \\
        Reptile+BN \cite{DBLP:journals/corr/abs-1803-02999-reptile} & BN & 49.9\textpm0.3 & 66.0\textpm0.6 & 32.0\textpm0.3 & 47.6\textpm0.3 & 52.4\textpm0.2 & 71.0\textpm0.2 & 35.3\textpm0.3 & 52.0\textpm0.3 \\
        ProtoNet \cite{Snell17_Proto} & No & 46.1\textpm0.8 & 65.8\textpm0.7 & 32.9\textpm0.5 & 49.3\textpm0.4 & 48.6\textpm0.9 & 69.6\textpm0.7 & 37.3\textpm0.6 & 57.8\textpm0.5 \\
        RelationNet \cite{sung2018learning} & BN &  51.4\textpm0.8 & 67.0\textpm0.7 & 34.9\textpm0.5 & 47.9\textpm0.4 & 54.5\textpm0.9 & 71.3\textpm0.8 & 36.3\textpm0.6 & 58.0\textpm0.6 \\
        TPN \cite{DBLP:journals/corr/abs-1805-10002-tpn} & Yes & 51.4\textpm0.8 & 67.1\textpm0.7 & 34.9\textpm0.5 & 47.9\textpm0.4 & 59.9\textpm0.9 & 73.3\textpm0.7 & 44.8\textpm0.6 & 59.4\textpm0.5 \\
        AttWeightGen \cite{DBLP:journals/corr/abs-1804-09458-dynamic} & BN & 56.2\textpm0.9 & 73.0\textpm0.6 & NA & NA & NA & NA & NA & NA\\
        TADAM \cite{NIPS2018_7352-tadam} & BN & 58.5\textpm0.3 & 76.7\textpm0.3 & NA & NA & NA & NA & NA & NA \\
        Simple CNAPS \cite{bateni2019improved} & BN & 53.2\textpm0.9 & 70.8\textpm0.7 & 37.1\textpm0.5 & 56.7\textpm0.5 & 63.0\textpm1.0 & 80.0\textpm0.8 & 48.1\textpm0.7 & 70.2\textpm0.6 \\
        LEO \cite{DBLP:journals/corr/abs-1807-05960-leo} & BN & 61.8\textpm0.1 & 77.6\textpm0.1 & NA & NA & 66.3\textpm0.1 & 81.4\textpm0.1 & NA & NA \\
        Transductive CNAPS & Yes & 55.6\textpm0.9 & 73.1\textpm0.7 & 42.8\textpm0.7 & 59.6\textpm0.5 & 65.9\textpm1.0 & 81.8\textpm0.7 & 54.6\textpm0.8 & 72.5\textpm0.6 \\
        \midrule
        Simple CNAPS \cite{bateni2019improved} + FETI & BN & 77.4\textpm0.8 & 90.3\textpm0.4 & 63.5\textpm0.6 & 83.1\textpm0.4 & 71.4\textpm1.0 & 86.0\textpm0.6 & 57.1\textpm0.7 & 78.5\textpm0.5 \\
        Transductive CNAPS + FETI & Yes & \textbf{79.9\textpm0.8} & \textbf{91.5\textpm0.4} & \textbf{68.5\textpm0.6} & \textbf{85.9\textpm0.3} & \textbf{73.8\textpm1.0} & \textbf{87.7\textpm0.6} & \textbf{65.1\textpm0.8} & \textbf{80.6\textpm0.5} \\
    \end{tabular}
    \vspace{-0.1in}
    \caption{Few-shot visual classification results on 1/5-shot 5/10-way few-shot on mini/tiered-ImageNet. For CNAP-based models, ``FETI'' indicates that the feature extractor used has been trained on ImageNet \cite{russakovsky2015imagenet} exluding classes within the test splits of mini/tiered-ImageNet (for more details see Appendix-\ref{appendix:benchmark-and-training:minitiered-training-and-testing}). ``BN'' indicates implicit transductive conditioning on the query set through the use of batch normalization. Error intervals denote 95\% confidence interval.}
    \vspace{-0.2in}
    \label{tab:results:mini-tiered-imagenet}
\end{table*}

\noindent
\textbf{Evaluation on Meta-Dataset: } In-domain, out-of-domain and overall rankings on Meta-Dataset are shown in Table \ref{tab:results:meta-dataset}. Following \cite{bateni2019improved} and \cite{requeima2019fast}, we pretrain the feature extractor on the training split of the ImageNet subset of Meta-Dataset. Transductive CNAPS sets new state of the art accuracy on 2 out of the 8 in-domain datasets, while matching other methods on 2 of the remaining domains. On out-of-domain tasks, it performs better with new state of the art performance on 4 out of the 5 out-of-domain datasets, Overall, it produces an average rank of 1.9 among all datasets, the best among the methods, with an average rank of 2.1 on in-domain tasks, only second to URT which was developed parallel to Transductive CNAPS, and 1.6 on out-of-domain tasks, the best among even the most recent methods. 

\vspace{0.02in}
\noindent
\textbf{Evaluation on mini/tiered-ImageNet: } We consider two feature extractor training settings on these benchmarks. First, we use the feature extractor trained on the corresponding training split of the mini/tiered-ImageNet. As shown in Table \ref{tab:results:mini-tiered-imagenet}, on tiered-ImageNet, Transductive CNAPS achieves state of art accuracy on both 10-way settings while matching state of the art performance of LEO \cite{DBLP:journals/corr/abs-1807-05960-leo} on the 5-way settings. On the mini-ImageNet, Transductive CNAPS out-performs other methods on 10-way settings while coming second to LEO \cite{DBLP:journals/corr/abs-1807-05960-leo} and TADAM \cite{requeima2019fast, NIPS2018_7352-tadam} on 5-way tasks. 

We attribute this difference in performance between mini-ImageNet and tiered-ImageNet to the fact that mini-ImageNet only provides 38,400 training examples, compared to 448,695 examples provided by tiered-ImageNet. This results in a lower performing ResNet-18 feature extractor (which is trained in a traditional supervised manner). This hypothesis is further supported by the results provided in our second model (denoted by ``FETI'', for ``Feature Extractor Trained with ImageNet'', in Table \ref{tab:results:mini-tiered-imagenet}). In this model, we train the feature extractor with a much larger subset of ImageNet, which has been carefully selected to prevent any possible overlap (in examples or classes) with the test sets of mini/tiered-ImageNet. Transductive CNAPS is able to take advantage of the more example-rich feature extractor, resulting in state-of-the-art performance across the board. Additionally, it outperforms the Simple CNAPS baseline by a large margin, even when using the same example-rich feature extractor; this demonstrates that leveraging additional query set information yields empirical gains.

\vspace{0.02in}
\noindent
\textbf{Performance vs. Class Shot: } In Figure \ref{fig:ratios-and-shots-all-ranges-10max}, we examine the relationship between class recall (i.e. accuracy among query examples belonging to the class itself) and the number of support examples in the class (shot). As shown, Transductive CNAPS is very effective when class shot is below 10, showing large average recall improvements, especially at the 1-shot level. However, as the class shot increases beyond 10, performance drops compared to Simple CNAPS. This suggests that soft k-means learning of cluster parameters can be effective when very few support examples are available. Conversely, in high-shot classes, transductive updates can act as distractors. %We also observe that the margin of improvement on low-\textit{shot} classes is diminished on out-of-domain tasks as compared to in-domain. This signifies the importance of properly adapting the feature space, a task that is more difficult to perform on out-of-domain tasks, and its impact on the effectiveness of our method.
\vspace{0.02in}

\begin{table*}[t]
    \centering
    \scriptsize
    \tabcolsep=0.09cm
    \begin{tabular}{lcccccccccccccccccc}
        & \multicolumn{8}{c}{In-Domain Accuracy (\%)} & \multicolumn{5}{c}{Out-of-Domain Accuracy (\%)} & \multicolumn{3}{c}{Avg Acc.} \\
        \cmidrule(lr){2-9}\cmidrule(lr){10-14}
        CNAPS Model & \rotatebox{45}{ImageNet} & \rotatebox{45}{Omniglot} & \rotatebox{45}{Aircraft} & \rotatebox{45}{Birds} & \rotatebox{45}{DTD} & \rotatebox{45}{QuickDraw} & \rotatebox{45}{Fungi} & \rotatebox{45}{Flower} & \rotatebox{45}{Signs} & \rotatebox{45}{MSCOCO} & \rotatebox{45}{MNIST} & \rotatebox{45}{CIFAR10} & \rotatebox{45}{CIFAR100} & In & Out & All \\
        \midrule
        GMM-EM+ & 53.3\textpm1.0 & 91.8\textpm0.6 & 81.2\textpm0.6 & \textbf{75.8\textpm0.7} & \textbf{71.8\textpm0.6} & 72.9\textpm0.7 & 42.8\textpm0.9 & 91.0\textpm0.4 & 66.1\textpm0.8 & 40.3\textpm1.0 & 94.2\textpm0.4 & 69.0\textpm0.7 & 51.3\textpm0.9 & 72.6 & 64.2 & 69.3 \\
        GMM & 45.3\textpm1.0 & 88.0\textpm0.9 & 80.8\textpm0.8 & 71.4\textpm0.8 & 61.1\textpm0.7 & 70.7\textpm0.8 & 42.9\textpm1.0 & 88.1\textpm0.6 & 68.9\textpm0.7 & 37.2\textpm0.9 & 91.4\textpm0.5 & 64.5\textpm0.7 & 46.6\textpm0.9 & 68.5 & 61.7 & 65.9 \\
        FEOT GMM & 52.6\textpm1.1 & 89.6\textpm0.7 & \textbf{84.0\textpm0.6} & 76.2\textpm0.8 & 66.5\textpm0.8 & 73.4\textpm0.8 & 45.7\textpm1.0 & 89.8\textpm0.6 & 74.4\textpm0.7 & 44.2\textpm1.0 & 93.1\textpm0.4 & 71.1\textpm0.8 & 56.9\textpm1.0 & 72.2 & 67.9 & 70.6 \\
        COT GMM & 48.7\textpm1.0 & 92.3\textpm0.5 & 80.0\textpm0.7 & 72.4\textpm0.7 & 59.8\textpm0.7 & 71.1\textpm0.7 & 41.4\textpm0.9 & 87.7\textpm0.5 & 63.6\textpm0.8 & 39.2\textpm0.8 & 89.8\textpm0.5 & 66.9\textpm0.7 & 50.5\textpm0.8 & 69.2 & 62.0 & 66.4 \\
        GMM-EM & 52.3\textpm1.0 & 92.0\textpm0.5 & \textbf{84.3\textpm0.6} & 75.2\textpm0.8 & 64.3\textpm0.7 & 72.6\textpm0.8 & 44.6\textpm1.0 & 90.8\textpm0.5 & 71.4\textpm0.7 & 44.7\textpm0.9 & 93.0\textpm0.4 & 71.1\textpm0.7 & 56.4\textpm0.9 & 72.0 & 67.3 & 70.2 \\
        \midrule
        Transductive+ & 53.3\textpm1.1 & 92.3\textpm0.5 & 81.2\textpm0.7 & 75.0\textpm0.8 & \textbf{72.0\textpm0.7} & 74.8\textpm0.8 & 45.1\textpm1.0 & 92.1\textpm0.4 & 71.0\textpm0.8 & 44.0\textpm1.1 & 95.9\textpm0.3 & 71.1\textpm0.7 & 57.3\textpm1.1 & 73.2 & 67.9 & 71.2 \\
        Simple & \textbf{58.6\textpm1.1} & 91.7\textpm0.6 & 82.4\textpm0.7 & 74.9\textpm0.8 & 67.8\textpm0.8 & 77.7\textpm0.7 & 46.9\textpm1.0 & 90.7\textpm0.5 & 73.5\textpm0.7 & 46.2\textpm1.1 & 93.9\textpm0.4 & 74.3\textpm0.7 & 60.5\textpm1.0 & 73.8 & 69.7 & 72.2 \\
        FEOT & \textbf{57.3\textpm1.1} & 90.5\textpm0.7 & 82.9\textpm0.7 & 74.8\textpm0.8 & 67.3\textpm0.8 & 76.3\textpm0.8 & 47.7\textpm1.0 & 90.5\textpm0.5 & \textbf{75.8\textpm0.7} & \textbf{47.1\textpm1.1} & 94.9\textpm0.4 & 74.3\textpm0.8 & \textbf{61.2\textpm1.0} & 73.4 & 70.7 & 72.4 \\
        COT & \textbf{58.8\textpm1.1} & \textbf{95.2\textpm0.3} & \textbf{84.0\textpm0.6} & \textbf{76.4\textpm0.7} & 68.5\textpm0.8 & \textbf{77.8\textpm0.7} & \textbf{49.7\textpm1.0} & \textbf{92.7\textpm0.4} & 70.8\textpm0.7 & \textbf{47.3\textpm1.0} & 94.2\textpm0.4 & \textbf{75.2\textpm0.7} & \textbf{61.2\textpm1.0} & \textbf{75.4} & 69.7 & \textbf{73.2} \\
        Transductive & \textbf{58.8\textpm1.1} & 93.9\textpm0.4 & \textbf{84.1\textpm0.6} & \textbf{76.8\textpm0.8} & 69.0\textpm0.8 & \textbf{78.6\textpm0.7} & \textbf{48.8\textpm1.1} & 91.6\textpm0.4 & \textbf{76.1\textpm0.7} & \textbf{48.7\textpm1.0} & \textbf{95.7\textpm0.3} & \textbf{75.7\textpm0.7} & \textbf{62.9\textpm1.0} & \textbf{75.2} & \textbf{71.8} & \textbf{73.9} \\
    \end{tabular}
    \vspace{-0.1in}
    \caption{Performance of various ablations of Tranductive and Simple CNAPS on Meta-Dataset. Error intervals indicate 95\% confidence intervals, and bold values indicate statistically significant state of the art performance.}
    \vspace{-0.1in}
    \label{tab:results:ablations}
\end{table*}{}

\begin{table*}
    \centering
    \scriptsize
    \tabcolsep=0.09cm
    \begin{tabular}{lcccccccccccccccc}
        & \multicolumn{8}{c}{In-Domain Accuracy (\%)} & \multicolumn{5}{c}{Out-of-Domain Accuracy (\%)} & \multicolumn{3}{c}{Avg Accuracy} \\
        \cmidrule(lr){2-9}\cmidrule(lr){10-14}
        Model & \rotatebox{45}{ImageNet} & \rotatebox{45}{Omniglot} & \rotatebox{45}{Aircraft} & \rotatebox{45}{Birds} & \rotatebox{45}{DTD} & \rotatebox{45}{QuickDraw} & \rotatebox{45}{Fungi} & \rotatebox{45}{Flower} & \rotatebox{45}{Signs} & \rotatebox{45}{MSCOCO} & \rotatebox{45}{MNIST} & \rotatebox{45}{CIFAR10} & \rotatebox{45}{CIFAR100} & In & Out & All \\
        \midrule
        Simple CNAPS & \textbf{58.6\textpm1.1} & 91.7\textpm0.6 & 82.4\textpm0.7 & 74.9\textpm0.8 & \textbf{67.8\textpm0.8} & \textbf{77.7\textpm0.7} & 46.9\textpm1.0 & 90.7\textpm0.5 & 73.5\textpm0.7 & 46.2\textpm1.1 & 93.9\textpm0.4 & 74.3\textpm0.7 & 60.5\textpm1.0 & 73.8 & 69.7 & 72.2 \\
        \midrule
        No Refinements & \textbf{57.3\textpm1.1} & 90.5\textpm0.7 & 82.9\textpm0.7 & 74.8\textpm0.8 & \textbf{67.3\textpm0.8} & 76.3\textpm0.8 & 47.7\textpm1.0 & 90.5\textpm0.5 & \textbf{75.8\textpm0.7} & \textbf{47.1\textpm1.1} & 94.9\textpm0.4 & 74.3\textpm0.8 & \textbf{61.2\textpm1.0} & 73.4 & 70.7 & 72.4 \\
        No Min/Max & \textbf{58.7\textpm1.1} & \textbf{94.0\textpm0.4} & \textbf{84.0\textpm0.6} & \textbf{76.4\textpm0.8} & \textbf{68.9\textpm0.8} & \textbf{77.9\textpm0.7} & 48.0\textpm1.0 & 91.6\textpm0.5 & 74.0\textpm0.8 & 48.3\textpm1.0 & 95.7\textpm0.3 & 75.5\textpm0.7 & 61.3\textpm1.0 & 74.9 & 71.0 & 73.4 \\
        Min 2 Max 4 & \textbf{58.8\textpm1.1} & \textbf{93.9\textpm0.4} & \textbf{84.1\textpm0.6} & \textbf{76.8\textpm0.8} & \textbf{69.0\textpm0.8} & \textbf{78.6\textpm0.7} & 48.8\textpm1.1 & \textbf{91.6\textpm0.4} & \textbf{76.1\textpm0.7} & \textbf{}48.7\textpm1.0 & \textbf{95.7\textpm0.3} & \textbf{75.7\textpm0.7} & \textbf{62.9\textpm1.0} & \textbf{75.2} & \textbf{71.8} & \textbf{73.9} \\
    \end{tabular}
    \vspace{-0.1in}
    \caption{Evaluating min/max refinement restrictions in Transductive CNAPS on Meta-Dataset, MNIST, and CIFAR10/100.}
    \vspace{-0.2in}
    \label{results:max-min-ablation}
\end{table*}{}

\vspace{0.02in}
\noindent
\textbf{Classification-Time Soft K-means Clustering:} We use soft k-means iterative updates of means and covariance at test-time only. It is natural to consider training the feature adaptation network end-to-end through the soft k-means transduction procedure. We provide this comparison in the bottom-half of Table \ref{tab:results:ablations}, with ``Transductive+ CNAPS'' denoting this variation. Iterative updates during training result in an average accuracy decrease of 2.5\%, which we conjecture to be due to training instabilities caused by applying this iterative algorithm early in training on noisy features.

%\vspace{0.02in}
\newpage
\noindent
\textbf{Transductive Feature Extraction vs. Classification:} \label{exp:feot-vs-cot} Our approach extends Simple CNAPS in two ways: improved adaptation of the feature extractor using a transductive task-encoding, and the soft k-means iterative estimation of class means and covariances.
We perform two ablations, ``Feature Extraction Only Transductive'' (FEOT) and ``Classification Only Transductive'' (COT), to independently assess the impact of these extensions.
 The results are presented in Table \ref{tab:results:ablations};  both extensions outperform Simple CNAPS. The transductive task-encoding is especially effective on out-of-domain tasks, whereas the soft k-mean learning of class parameters boosts accuracy on in-domain tasks. Transductive CNAPS is able to leverage the best of both worlds, allowing it to achieve significant gains over Simple CNAPS.

\vspace{0.02in}
\noindent
\textbf{Comparison to Gaussian Mixture Models:} \label{exp:gaussian-mixture-models} 
We consider five GMM-based ablations of our method where the log-determinant is introduced into the weight updates (using a uniform class prior $\pi_k = 1/K$). Results in Table \ref{tab:results:ablations} correspond to their soft k-means counterparts in the same order shown. The GMM-based variations of our method and Simple CNAPS result in a notable 4-8\% loss in overall accuracy. It is also surprising to observe that the FEOT variation matches the performance of the full GMM-EM model.

\vspace{0.02in}
\noindent
\textbf{Maximum and Minimum Number of Refinements:} In our experiments, we use a minimum number of 2 refinement steps of class parameters, with the maximum set to 4 on the Meta-Dataset and 10 on the mini/tiered-ImageNet benchmarks. As shown in Table \ref{results:max-min-ablation}, the refinement criteria itself, without any step constraints, results in a significant performance gain as compared to performing no refinements. In fact, it accounts for the majority of the accuracy gain for Transductive CNAPS. We further explore the impact of these step-hyperparameters on the performance on Transductive CNAPS on the Meta-Dataset in Figure \ref{fig:max-and-min}. As shown, requiring the same number of refinement steps for every task results in sub-optimal performance. This is demonstrated by the fact that the peak performance for each minimum number of steps is achieved with a larger number of maximum steps, showcasing the importance of allowing different numbers of refinement steps depending on the task. In addition, we observe that as the number of minimum refinement steps increases, the performance improves up to two steps while declining after. This suggests that, unlike \cite{DBLP:journals/corr/abs-1803-00676-tieredimagenet} where only a single refinement step leads to the best performance, our Mahalanobis-based approach can leverage extra steps to further refine the class parameters. We do see a decline in performance with a higher number of steps; this suggests that while our refinement criteria can be effective at performing different number of steps depending on the task, it can potentially lead to over-fitting, justifying the need for a well chosen maximum number of steps.

\section{Discussion}
In this paper, we have presented a few-shot visual classification method that achieves new state of the art performance via a transductive clustering procedure for refining class parameters derived from a previous neural adaptive Mahalanobis-distance based approach. 
The resulting architecture, Transductive CNAPS, is more effective at producing useful estimates of class mean and covariance especially in low-shot settings, when used at test time. Even though we demonstrate the efficacy of our approach in the transductive domain where query examples themselves are used as unlabelled data, our soft k-means clustering procedure naturally extends to use other sources of unlabelled examples in a semi-supervised fashion. 

Transductive CNAPS superficially resembles a transductive GMM stacked on top of a learned feature representation; however, when we try to make this connection exact (by including the log-determinant of the class covariances), we suffer substantial performance hits. Explaining why this happens will be the subject of future work.
% This suggests that the appropriate framework to reason about the Transductive CNAPS and Simple CNAPS family of models is through local metric learning \cite{ramanan2010local, weinberger2009distance, hauberg2012geometric}; the class precision matrices $\mathbf Q_k^{-1}$ play the role of the Riemannian metric at the class centroids, and the resulting classifier is a coarse approximation of a distance-based classifier on the induced Riemannian manifold. We discuss this connection in more depth in Appendix \ref{appendix:differential-geometry}. Working out the exact nature of the geometric connection, and why the probabilistic model fails empirically, is a subject for future research. 

\section{Acknowledgments}
We acknowledge the support of the Natural Sciences and Engineering Research Council of Canada (NSERC), the Canada Research Chairs (CRC) Program, the Canada CIFAR AI Chairs Program, Compute Canada, Intel, Inverted AI under their MITAC Accelerate internship program and DARPA under its D3M and LWLL programs. Additionally, this material is based upon work supported by the United States Air Force under Contract No.
FA8750-19-C-0515.

{\small
\bibliographystyle{ieee_fullname}
\bibliography{main}

\begin{thebibliography}{10}\itemsep=-1pt

\bibitem{banerjee2005clustering}
Arindam Banerjee, Srujana Merugu, Inderjit~S Dhillon, and Joydeep Ghosh.
\newblock Clustering with bregman divergences.
\newblock {\em Journal of Machine Learning Research}, 6(Oct):1705--1749, 2005.

\bibitem{bateni2019improved}
Peyman Bateni, Raghav Goyal, Vaden Masrani, Frank Wood, and Leonid Sigal.
\newblock Improved few-shot visual classification.
\newblock In {\em IEEE/CVF Conference on Computer Vision and Pattern
  Recognition (CVPR)}, 2020.

\bibitem{bellet2013survey}
Aur{\'e}lien Bellet, Amaury Habrard, and Marc Sebban.
\newblock A survey on metric learning for feature vectors and structured data.
\newblock {\em arXiv preprint arXiv:1306.6709}, 2013.

\bibitem{cimpoi2014describing}
Mircea Cimpoi, Subhransu Maji, Iasonas Kokkinos, Sammy Mohamed, and Andrea
  Vedaldi.
\newblock Describing textures in the wild.
\newblock In {\em IEEE/CVF Conference on Computer Vision and Pattern
  Recognition (CVPR)}, pages 3606--3613, 2014.

\bibitem{dvornik2020selecting-sur}
Nikita Dvornik, Cordelia Schmid, and Julien Mairal.
\newblock Selecting relevant features from a multi-domain representation for
  few-shot classification, 2020.

\bibitem{feyjie2020semisupervised-medical}
Abdur~R Feyjie, Reza Azad, Marco Pedersoli, Claude Kauffman, Ismail~Ben Ayed,
  and Jose Dolz.
\newblock Semi-supervised few-shot learning for medical image segmentation,
  2020.

\bibitem{finn2017model}
Chelsea Finn, Pieter Abbeel, and Sergey Levine.
\newblock Model-agnostic meta-learning for fast adaptation of deep networks.
\newblock In {\em International Conference on Machine Learning (ICML)}, 2017.

\bibitem{DBLP:journals/corr/abs-1807-01613-cnp}
Marta Garnelo, Dan Rosenbaum, Chris~J. Maddison, Tiago Ramalho, David Saxton,
  Murray Shanahan, Yee~Whye Teh, Danilo~J. Rezende, and S.~M.~Ali Eslami.
\newblock Conditional neural processes.
\newblock {\em International Conference on Machine Learning (ICML)}, 2018.

\bibitem{DBLP:journals/corr/abs-1804-09458-dynamic}
Spyros Gidaris and Nikos Komodakis.
\newblock Dynamic few-shot visual learning without forgetting.
\newblock {\em IEEE/CVF Conference on Computer Vision and Pattern Recognition
  (CVPR)}, abs/1804.09458, 2018.

\bibitem{Goodfellow2014_GANS}
Ian Goodfellow, Jean Pouget-Abadie, Mehdi Mirza, Bing Xu, David Warde-Farley,
  Sherjil Ozair, Aaron Courville, and Yoshua Bengio.
\newblock Generative adversarial nets.
\newblock In {\em Advances in Neural Information Processing Systems},
  volume~27, pages 2672--2680, 2014.

\bibitem{guz-etal-2020-neural}
Grigorii Guz, Peyman Bateni, Darius Muglich, and Giuseppe Carenini.
\newblock Neural {RST}-based evaluation of discourse coherence.
\newblock In {\em Proceedings of the 1st Conference of the Asia-Pacific Chapter
  of the Association for Computational Linguistics and the 10th International
  Joint Conference on Natural Language Processing}, pages 664--671, 2020.

\bibitem{He15_ResNet}
K. {He}, X. {Zhang}, S. {Ren}, and J. {Sun}.
\newblock Deep residual learning for image recognition.
\newblock In {\em IEEE/CVF Conference on Computer Vision and Pattern
  Recognition (CVPR)}, 2016.

\bibitem{DBLP:journals/corr/HeZRS15-resnet}
Kaiming He, Xiangyu Zhang, Shaoqing Ren, and Jian Sun.
\newblock Deep residual learning for image recognition.
\newblock {\em IEEE/CVF Conference on Computer Vision and Pattern Recognition
  (CVPR)}, 2016.

\bibitem{Hossain:2019:CSD:3303862.3295748-image-captioning-survey}
MD.~Zakir Hossain, Ferdous Sohel, Mohd~Fairuz Shiratuddin, and Hamid Laga.
\newblock A comprehensive survey of deep learning for image captioning.
\newblock {\em ACM Comput. Surv.}, 51(6), Feb. 2019.

\bibitem{houben2013detection}
Sebastian Houben, Johannes Stallkamp, Jan Salmen, Marc Schlipsing, and
  Christian Igel.
\newblock Detection of traffic signs in real-world images: The german traffic
  sign detection benchmark.
\newblock In {\em International Joint Conf. on Neural Networks (IJCNN)}, pages
  1--8, 2013.

\bibitem{DBLP:journals/corr/abs-1907-09408-object-detection-survey}
Licheng Jiao, Fan Zhang, Fang Liu, Shuyuan Yang, Lingling Li, Zhixi Feng, and
  Rong Qu.
\newblock A survey of deep learning-based object detection.
\newblock {\em CoRR}, abs/1907.09408, 2019.

\bibitem{jongejan2016quick}
Jonas Jongejan, Henry Rowley, Takashi Kawashima, Jongmin Kim, and Nick
  Fox-Gieg.
\newblock The quick, draw!-ai experiment.(2016), 2016.

\bibitem{DBLP:journals/corr/abs-1905-01436-edge-labelling-gnn}
Jongmin Kim, Taesup Kim, Sungwoong Kim, and Chang~D. Yoo.
\newblock Edge-labeling graph neural network for few-shot learning.
\newblock {\em IEEE/CVF Conference on Computer Vision and Pattern Recognition
  (CVPR)}, 2019.

\bibitem{koch2015siamese}
Gregory Koch, Richard Zemel, and Ruslan Salakhutdinov.
\newblock Siamese neural networks for one-shot image recognition.
\newblock In {\em ICML deep learning workshop}, volume~2, 2015.

\bibitem{Krizhevsky09learningmultiple}
Alex Krizhevsky.
\newblock Learning multiple layers of features from tiny images.
\newblock Technical report, 2009.

\bibitem{Krizhevsky12_AlexNet}
Alex Krizhevsky, Ilya Sutskever, and Geoffrey Hinton.
\newblock Imagenet classification with deep convolutional neural networks.
\newblock {\em Neural Information Processing Systems}, 25, 2012.

\bibitem{Krizhevsky:2017:ICD:3098997.3065386alexnet}
Alex Krizhevsky, Ilya Sutskever, and Geoffrey~E. Hinton.
\newblock Imagenet classification with deep convolutional neural networks.
\newblock {\em Commun. ACM}, 60(6):84--90, May 2017.

\bibitem{lake2015human}
Brenden~M Lake, Ruslan Salakhutdinov, and Joshua~B Tenenbaum.
\newblock Human-level concept learning through probabilistic program induction.
\newblock {\em Science}, 350(6266):1332--1338, 2015.

\bibitem{lecun-mnisthandwrittendigit-2010}
Yann LeCun and Corinna Cortes.
\newblock {MNIST} handwritten digit database.
\newblock 2010.

\bibitem{Lichtenstein2020_TAFSSL}
Moshe Lichtenstein, Prasanna Sattigeri, Rog{\'{e}}rio Feris, Raja Giryes, and
  Leonid Karlinsky.
\newblock {TAFSSL:} task-adaptive feature sub-space learning for few-shot
  classification.
\newblock In Andrea Vedaldi, Horst Bischof, Thomas Brox, and Jan{-}Michael
  Frahm, editors, {\em Computer Vision - {ECCV} 2020 - 16th European
  Conference, Glasgow, UK, August 23-28, 2020, Proceedings, Part {VII}}, volume
  12352 of {\em Lecture Notes in Computer Science}, pages 522--539. Springer,
  2020.

\bibitem{lin2014microsoft}
Tsung-Yi Lin, Michael Maire, Serge Belongie, James Hays, Pietro Perona, Deva
  Ramanan, Piotr Doll{\'a}r, and C~Lawrence Zitnick.
\newblock Microsoft coco: Common objects in context.
\newblock In {\em European Conference on Computer Vision (ECCV)}, pages
  740--755, 2014.

\bibitem{liu2020universal-urt}
Lu Liu, William Hamilton, Guodong Long, Jing Jiang, and Hugo Larochelle.
\newblock A universal representation transformer layer for few-shot image
  classification, 2020.

\bibitem{DBLP:journals/corr/abs-1805-10002-tpn}
Yanbin Liu, Juho Lee, Minseop Park, Saehoon Kim, and Yi Yang.
\newblock Learning to propagate labels: Transductive propagation network for
  few-shot learning.
\newblock {\em International Conference on Learning Representations (ICLR)},
  2019.

\bibitem{maji2013fine}
Subhransu Maji, Esa Rahtu, Juho Kannala, Matthew Blaschko, and Andrea Vedaldi.
\newblock Fine-grained visual classification of aircraft.
\newblock {\em arXiv preprint arXiv:1306.5151}, 2013.

\bibitem{melnykov2014k}
Igor Melnykov and Volodymyr Melnykov.
\newblock On k-means algorithm with the use of mahalanobis distances.
\newblock {\em Statistics \& Probability Letters}, 84:88--95, 2014.

\bibitem{DBLP:journals/corr/MishraRCA17-snail}
Nikhil Mishra, Mostafa Rohaninejad, Xi Chen, and Pieter Abbeel.
\newblock Meta-learning with temporal convolutions.
\newblock {\em CoRR}, abs/1707.03141, 2017.

\bibitem{DBLP:journals/corr/abs-1803-02999-reptile}
Alex Nichol, Joshua Achiam, and John Schulman.
\newblock On first-order meta-learning algorithms.
\newblock {\em CoRR}, abs/1803.02999, 2018.

\bibitem{nilsback2008automated}
Maria-Elena Nilsback and Andrew Zisserman.
\newblock Automated flower classification over a large number of classes.
\newblock In {\em IEEE Indian Conference on Computer Vision, Graphics \& Image
  Processing}, pages 722--729, 2008.

\bibitem{NIPS2018_7352-tadam}
Boris Oreshkin, Pau Rodr\'{\i}guez~L\'{o}pez, and Alexandre Lacoste.
\newblock Tadam: Task dependent adaptive metric for improved few-shot learning.
\newblock In {\em Advances in Neural Information Processing Systems}, pages
  721--731. 2018.

\bibitem{perez2018film}
Ethan Perez, Florian Strub, Harm De~Vries, Vincent Dumoulin, and Aaron
  Courville.
\newblock Film: Visual reasoning with a general conditioning layer.
\newblock In {\em Thirty-Second AAAI Conference on Artificial Intelligence},
  2018.

\bibitem{Qiao_2019_ICCV-team}
Limeng Qiao, Yemin Shi, Jia Li, Yaowei Wang, Tiejun Huang, and Yonghong Tian.
\newblock Transductive episodic-wise adaptive metric for few-shot learning.
\newblock In {\em IEEE/CVF International Conference on Computer Vision (ICCV)},
  2019.

\bibitem{DBLP:conf/iclr/RaviL17-meta-lstm}
Sachin Ravi and Hugo Larochelle.
\newblock Optimization as a model for few-shot learning.
\newblock In {\em International Conference on Learning Representations (ICLR)},
  2017.

\bibitem{Redmon16_YOLO}
J. {Redmon}, S. {Divvala}, R. {Girshick}, and A. {Farhadi}.
\newblock You only look once: Unified, real-time object detection.
\newblock In {\em IEEE/CVF Conference on Computer Vision and Pattern
  Recognition (CVPR)}, 2016.

\bibitem{DBLP:journals/corr/abs-1803-00676-tieredimagenet}
Mengye Ren, Eleni Triantafillou, Sachin Ravi, Jake Snell, Kevin Swersky,
  Joshua~B. Tenenbaum, Hugo Larochelle, and Richard~S. Zemel.
\newblock Meta-learning for semi-supervised few-shot classification.
\newblock {\em International Conference on Learning Representations (ICLR)},
  2018.

\bibitem{Ren15_FasterRCNN}
Shaoqing Ren, Kaiming He, Ross Girshick, and Jian Sun.
\newblock Faster {R-CNN}: Towards real-time object detection with region
  proposal networks.
\newblock {\em IEEE Transactions on Pattern Analysis and Machine Intelligence
  (TPAMI)}, 39, June 2015.

\bibitem{requeima2019fast}
James Requeima, Jonathan Gordon, John Bronskill, Sebastian Nowozin, and
  Richard~E Turner.
\newblock Fast and flexible multi-task classification using conditional neural
  adaptive processes.
\newblock {\em Advances in Neural Information Processing Systems}, 2019.

\bibitem{russakovsky2015imagenet}
Olga Russakovsky, Jia Deng, Hao Su, Jonathan Krause, Sanjeev Satheesh, Sean Ma,
  Zhiheng Huang, Andrej Karpathy, Aditya Khosla, Michael Bernstein, et~al.
\newblock Imagenet large scale visual recognition challenge.
\newblock {\em International Journal of Computer Vision (IJCV)},
  115(3):211--252, 2015.

\bibitem{DBLP:journals/corr/abs-1807-05960-leo}
Andrei~A. Rusu, Dushyant Rao, Jakub Sygnowski, Oriol Vinyals, Razvan Pascanu,
  Simon Osindero, and Raia Hadsell.
\newblock Meta-learning with latent embedding optimization.
\newblock {\em International Conference on Learning Representations (ICLR)},
  2018.

\bibitem{garcia2018fewshot}
Victor~Garcia Satorras and Joan~Bruna Estrach.
\newblock Few-shot learning with graph neural networks.
\newblock In {\em International Conference on Learning Representations (ICLR)},
  2018.

\bibitem{fungi2018schroeder}
B. Schroeder and Y Cui.
\newblock Fgvcx fungi classification challenge 2018, 2018.

\bibitem{Scibior2021_ITRA}
Adam Scibior, Vasileios Lioutas, Daniele Reda, Peyman Bateni, and Frank Wood.
\newblock Imagining the road ahead: Multi-agent trajectory prediction via
  differentiable simulation, 2021.

\bibitem{vgg-paper-2014}
Karen Simonyan and Andrew Zisserman.
\newblock Very deep convolutional networks for large-scale image recognition.
\newblock {\em arXiv 1409.1556}, 09 2014.

\bibitem{Snell17_Proto}
Jake Snell, Kevin Swersky, and Richard Zemel.
\newblock Prototypical networks for few-shot learning.
\newblock In {\em Advances in Neural Information Processing Systems}, 2017.

\bibitem{Song_2020_CVPR-cross-attention}
Xibin Song, Yuchao Dai, Dingfu Zhou, Liu Liu, Wei Li, Hongdong Li, and Ruigang
  Yang.
\newblock Channel attention based iterative residual learning for depth map
  super-resolution.
\newblock In {\em IEEE/CVF Conference on Computer Vision and Pattern
  Recognition (CVPR)}, 2020.

\bibitem{8441512-image-classification-survey}
M. {Sornam}, K. {Muthusubash}, and V. {Vanitha}.
\newblock A survey on image classification and activity recognition using deep
  convolutional neural network architecture.
\newblock In {\em International Conference on Advanced Computing (ICoAC)},
  2017.

\bibitem{sung2018learning}
Flood Sung, Yongxin Yang, Li Zhang, Tao Xiang, Philip~HS Torr, and Timothy~M
  Hospedales.
\newblock Learning to compare: Relation network for few-shot learning.
\newblock In {\em IEEE/CVF Conference on Computer Vision and Pattern
  Recognition (CVPR)}, 2018.

\bibitem{DBLP:journals/corr/SzegedyLJSRAEVR14-inception}
Christian Szegedy, Wei Liu, Yangqing Jia, Pierre Sermanet, Scott~E. Reed,
  Dragomir Anguelov, Dumitru Erhan, Vincent Vanhoucke, and Andrew Rabinovich.
\newblock Going deeper with convolutions.
\newblock {\em CoRR}, abs/1409.4842, 2014.

\bibitem{triantafillou2019meta}
Eleni Triantafillou, Tyler Zhu, Vincent Dumoulin, Pascal Lamblin, Kelvin Xu,
  Ross Goroshin, Carles Gelada, Kevin Swersky, Pierre-Antoine Manzagol, and
  Hugo Larochelle.
\newblock Meta-dataset: A dataset of datasets for learning to learn from few
  examples.
\newblock {\em International Conference on Learning Representations (ICLR)},
  2020.

\bibitem{vinyals2016matching}
Oriol Vinyals, Charles Blundell, Timothy Lillicrap, Daan Wierstra, et~al.
\newblock Matching networks for one shot learning.
\newblock In {\em Advances in neural information processing systems}, pages
  3630--3638, 2016.

\bibitem{wah2011caltech}
Catherine Wah, Steve Branson, Peter Welinder, Pietro Perona, and Serge
  Belongie.
\newblock The caltech-ucsd birds-200-2011 dataset.
\newblock 2011.

\bibitem{Wang:2019:SZL:3306498.3293318-survey-of-zero-shot-learning}
Wei Wang, Vincent~W. Zheng, Han Yu, and Chunyan Miao.
\newblock A survey of zero-shot learning: Settings, methods, and applications.
\newblock {\em ACM Trans. Intell. Syst. Technol.}, 10(2):13:1--13:37, Jan.
  2019.

\bibitem{DBLP:journals/corr/abs-1904-05046-survey-on-few-shot-learning}
Yaqing Wang and Quanming Yao.
\newblock Few-shot learning: {A} survey.
\newblock {\em CoRR}, abs/1904.05046, 2019.

\bibitem{DBLP:journals/corr/YosinskiCBL14-finetune}
Jason Yosinski, Jeff Clune, Yoshua Bengio, and Hod Lipson.
\newblock How transferable are features in deep neural networks?
\newblock {\em Advances in Neural Information Processing Systems}, 2014.

\bibitem{Zhao2021_ReMP}
Yang Zhao, Chunyuan Li, Ping Yu, and Changyou Chen.
\newblock Remp: Rectified metric propagation for few-shot learning.
\newblock In {\em Proceedings of the IEEE/CVF Conference on Computer Vision and
  Pattern Recognition (CVPR) Workshops}, pages 2581--2590, June 2021.

\end{thebibliography}
}

\newpage
\clearpage

\begin{appendices}

\section{Benchmarks and Training}
\label{appendix:benchmark-and-training}

\subsection{Meta-Dataset}
\label{appendix:benchmark-and-training:meta-dataset}

\begin{figure}[b]
    \centering
    \small
    \subfloat[Number of Tasks vs. Ways]{\label{fig:ways-frequency}{\includegraphics[width=3.2765in]{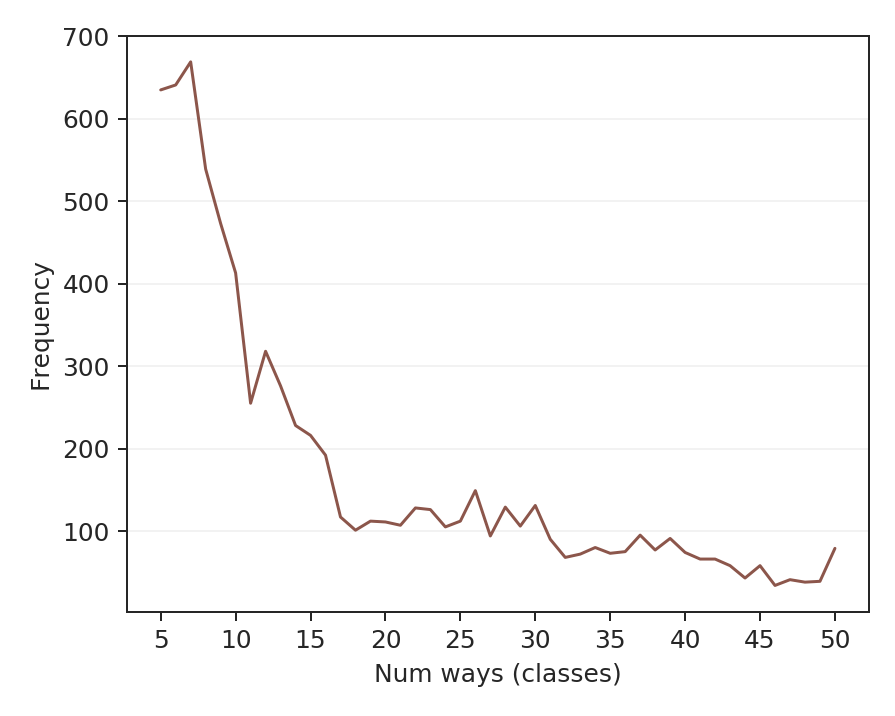} \label{fig:ways-vs-freq}}} \\
    \vspace{-0.1in}
    \subfloat[Number of Classes vs. Shots]{\label{fig:shot-frequency}{\includegraphics[width=3.2765in]{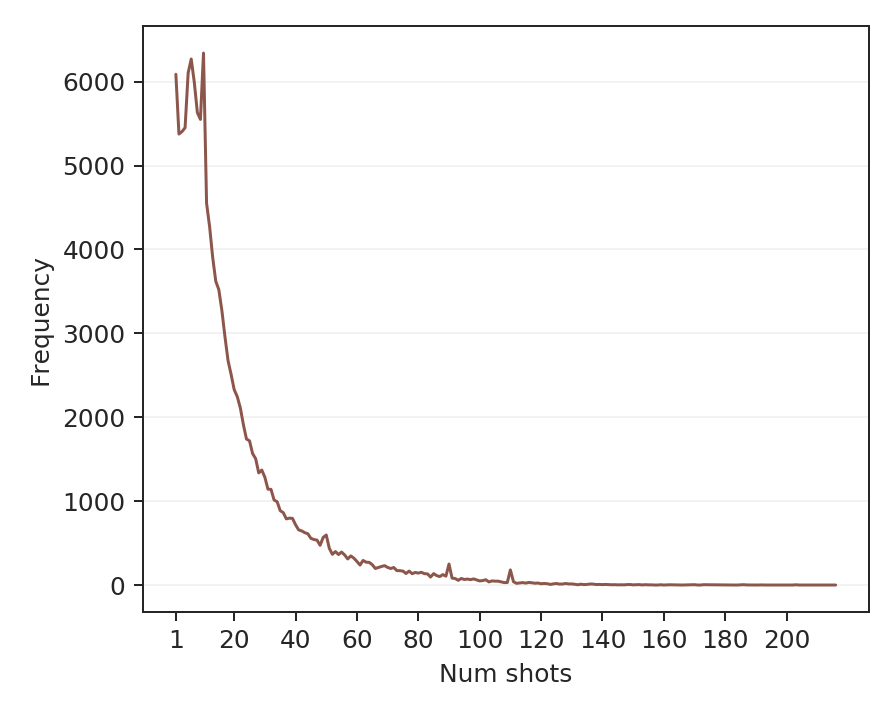} \label{fig:shots-vs-freq}}}%
    \vspace{-0.1in}
    \caption{Test-time way and shot frequency graphs. Figure is directly from Bateni et al. \cite{bateni2019improved}. As shown, most tasks have fewer than 10 classes (way) and most classes have less than 20 support examples (shot).}
    \vspace{-0.1in}
\end{figure}

A brief description of the sampling procedure used in the Meta-Dataset setting is already provided in Section \ref{experiments:benchmarks}.
This sampling procedure, however, comes with additional specifications that are uniform across all tasks (such as count enforcing) and dataset specific details such as considering the class hierarchy in ImageNet tasks. 
The full algorithm for sampling is outlined in \cite{triantafillou2019meta}, and we refer the interested reader to Section 3.2 in \cite{triantafillou2019meta} for complete details. This procedure results in a task distribution where most tasks have fewer than 10 classes and each class has fewer than 20 support examples. The task frequency relative to the number of classes is presented in Figure \ref{fig:ways-frequency}, and the class frequency as compared to the class shot is presented in Figure \ref{fig:shot-frequency}. The query set contains between 1 and 10 (inclusive) examples per class for all tasks; fewer than 10 query examples occur only when there are not enough total images to support 10 query examples.
% As for query examples, across all tasks and classes, 10 query examples are used per class in the task and evaluate classification accuracy, and additionally, as is the case with our method, are used for transductive learning. Note that 10 query examples are sampled only when 10 examples are left for the class after sampling of the support set. If not, the maximum number of images left is chosen with minimum of 1 query image per class.

\subsection{mini/tiered-ImageNet}
\label{appendix:benchmark-and-training:minitiered}
Task sampling across both mini-ImageNet and tiered-ImageNet first starts by defining a constant number of ways and shots that will be used for each generated task. For a $L$-shot $K$-way problem setting, first $K$ classes are sampled from the dataset with uniform probability. Then, for each sampled class, $L$ of the class images are sampled with uniform probability and used as the support examples for the class. In addition, 10 query images (distinct from the support images) are sampled per class.% where a total of $10 \times K$ query examples are present in the class.

\subsection{Meta-Dataset Training/Testing}
\label{appendix:benchmark-and-training:meta-dataset-training-and-testing}

Following \cite{bateni2019improved} and \cite{requeima2019fast}, we train our ResNet18 feature extractor as a supervised multi-class image classifier on the training split of the ImageNet subset of the Meta-Dataset. We directly follow the procedure described in C.1.1 of \cite{requeima2019fast} and use their released checkpoint. Images from the 712 ImageNet classes designated for training by Meta-Dataset \cite{triantafillou2019meta} are first resized to 84x84. The ResNet18 in Transductive CNAPS is then trained as a 712-class image recognition task using this data. Training is done using the cross-entropy loss for 125
epochs using SGD with momentum of 0.9, weight decay of 0.0001,
batch size of 256, and a learning rate of 0.1 that is reduced by a factor of 10 every 25 epochs.
During training, the dataset was augmented with random crops, horizontal flips,
and color jitter.

Once the ResNet18 is trained, we freeze the parameters and proceed to train the adaptation network using Episodic training \cite{Snell17_Proto, finn2017model} where tasks themselves are used as training examples.
For each iteration of Episodic training, a task (with additional ground truth query labels) is generated, and the adaptation network is trained to minimize classification error (cross entropy) of the query set given the task.
% In Episodic training, every training iteration (episode), a task is provided where the network is adapted using the support examples and class probabilities are produced for query examples. The ground truth labels for the query examples are then used to calculate a cross-entropy loss on the produced class probabilities. This is used for backpropagation across the network with the exception of the ResNet18 feature extractor that has been pretrained and is fixed.
We train for a total of 110K tasks, with 16 tasks per batch, resulting in 6875 gradient updates. We train using Adam optimizer with learning rate of $5 \times 10^{-4}$. We evaluate on the validation splits of all 8 in-domain and 1 out-of-domain (MSCOCO) datasets, saving the best performing checkpoint for test-time evaluation.

\subsection{mini/tiered-ImageNet Training/Testing}
\label{appendix:benchmark-and-training:minitiered-training-and-testing}
Similar to Meta-Dataset, we first train the ResNet18 feature extractor. This is done with respect to two settings: first, we directly use the training data from mini-ImageNet and tiered-ImageNet, with 38,400 and 448,695 images respectively. Second, we consider a larger training set of 825 classes from ImageNet \cite{russakovsky2015imagenet} that don't overlap with the test sets of the benchmarks. In both cases, we follow the same procedure as the one described for Meta-Dataset in \ref{appendix:benchmark-and-training:meta-dataset-training-and-testing} with the exception of training for 90 epochs and reducing the learning rate every 30 epochs. 

After training the ResNet18, the weights are frozen while we train the adaptation network using Episodic training: at each iteration, a task is generated, and we backpropagate the query set classification loss through the adaptation network. For mini/tiered-ImageNet, we train for a total of 20K tasks, validating performance every 2K tasks and saving the best checkpoint for test-time evaluation. We, similarly, use the Adam optimiser with learning rate of $5 \times 10^{-4}$, and use a batch size of 16, for a total of 1250 gradient steps.

\begin{figure}[t]
    \centering
    \small
    \vspace{-0.1in}
    \subfloat[5-way 1-shot]{\label{fig:query-num-vs-performance}{\includegraphics[width=0.48\textwidth]{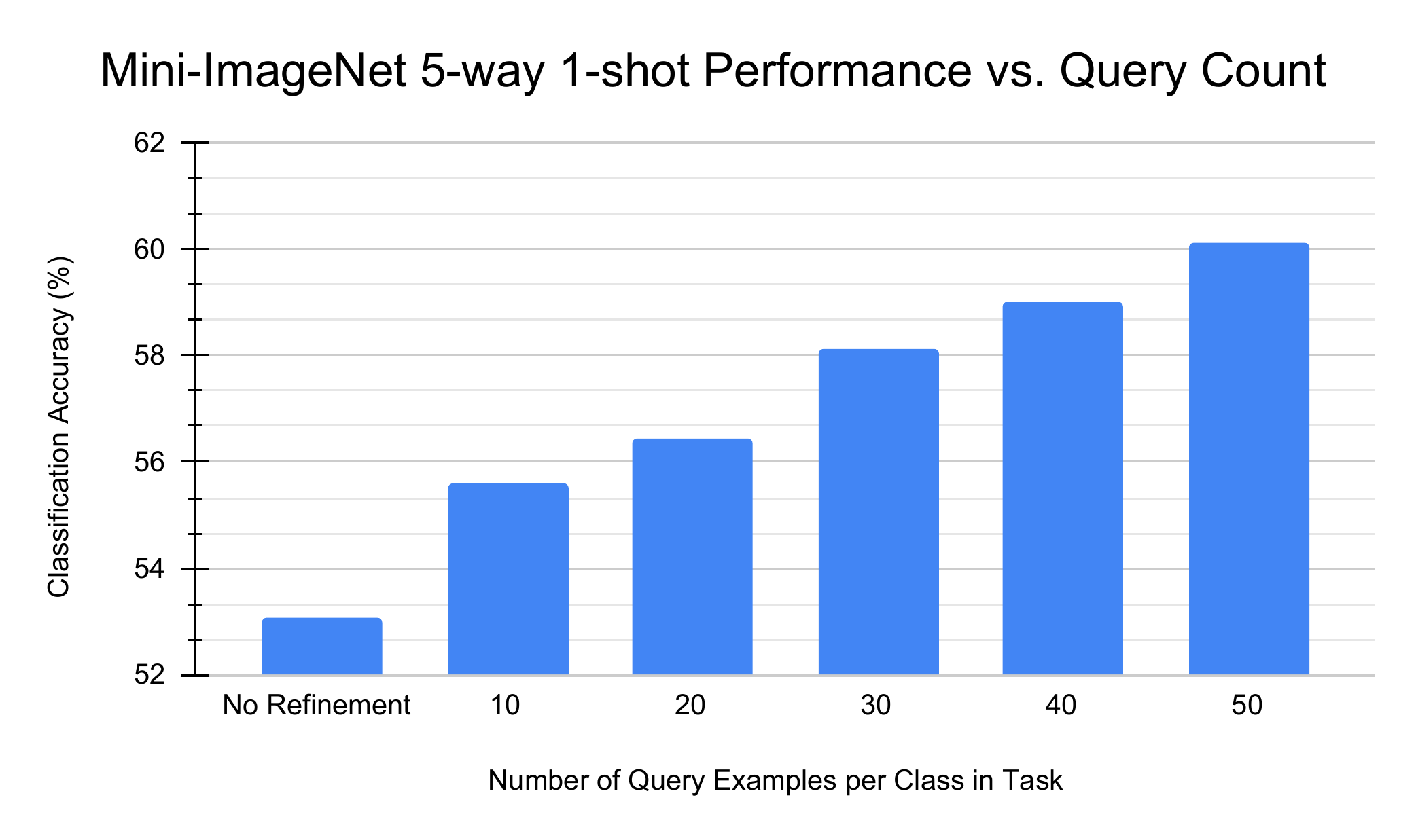}}}\\
    \vspace{-0.05in}
    \subfloat[5-way 5-shot]{\label{fig:query-num-vs-performance}{\includegraphics[width=0.48\textwidth]{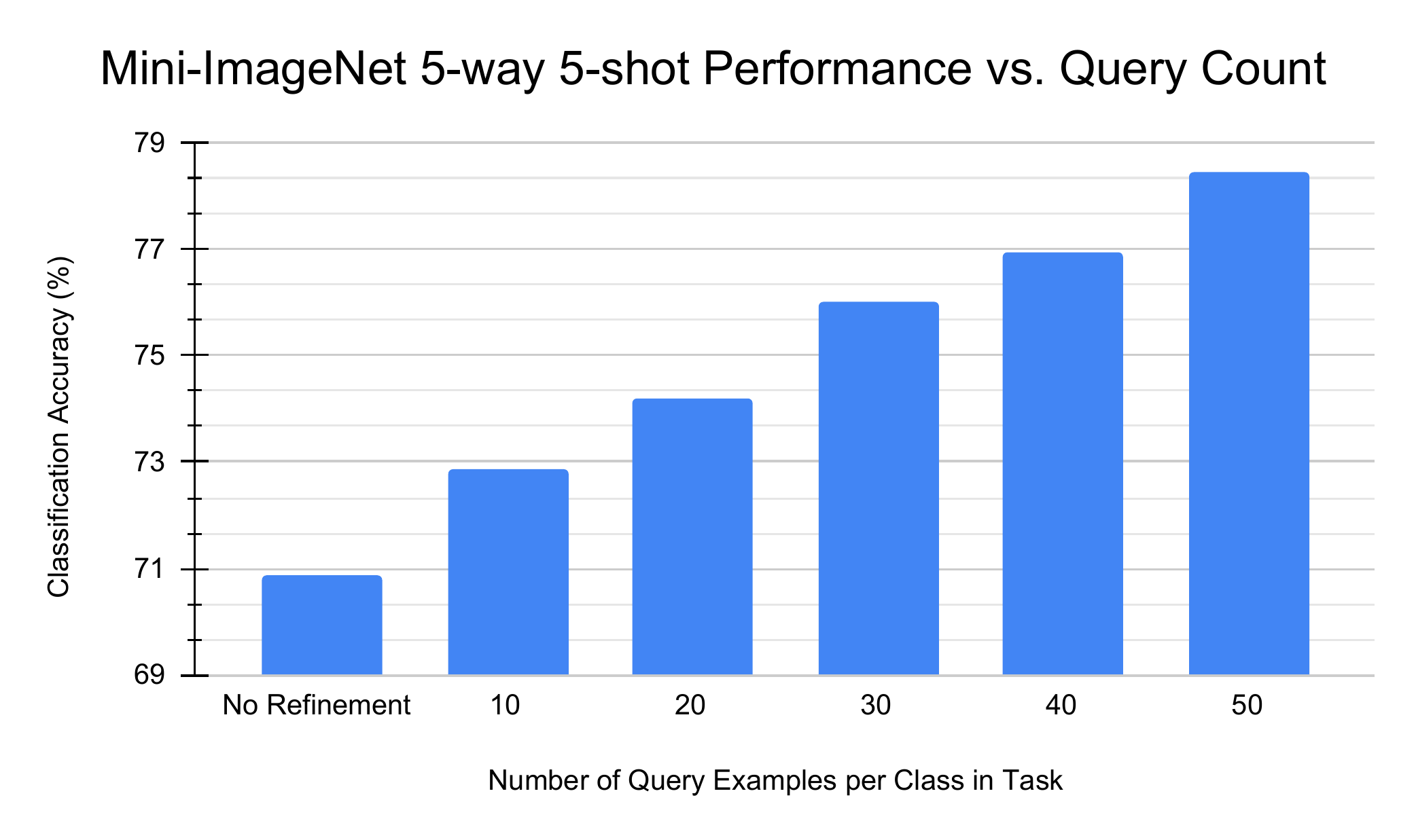}}}%
    \vspace{-0.05in}
    \caption{Mini-ImageNet performance as the number of query examples per class increases on 1/5-shot 5-way tasks.}
    \label{fig:query-num-vs-performance}
    \vspace{-0.2in}
\end{figure}

\section{Additional Experiments}

\subsection{Performance vs. Number of Query Examples}

Figure \ref{fig:query-num-vs-performance} shows the performance of Transductive CNAPS on 1/5-shot 5-way Mini-ImageNet tasks as the number of query examples per category increases from 10 to 50. As shown, with a greater number of query examples, performance improves as our method is able to exploit more unlabelled data.

\end{appendices}

\end{document}